\documentclass{article}

\usepackage[T2A]{fontenc}
\usepackage[utf8]{inputenc}
\usepackage[english]{babel}
\usepackage{amsmath}
\usepackage{amssymb}
\usepackage{amsthm}
\usepackage{mathrsfs}

\usepackage{dan2e}

\usepackage{subfig}

\theoremstyle{definition}
\newtheorem{definition}{Definition}
\theoremstyle{plain}
\newtheorem{remark}{Remark}
\newtheorem{theorem}{Theorem}

\newtheorem{proposition}[theorem]{Proposition}
\usepackage{url}

\usepackage{hyperref}       
\usepackage{booktabs}       
\usepackage{amsfonts}       
\usepackage{nicefrac}       
\usepackage{microtype}      
\usepackage{bm}

\usepackage[algoruled,boxed,lined]{algorithm2e}

\usepackage{graphicx}
\begin{document}

\Volume{514:2}
\Year{2023}
\Pages{196-211}

\udk{}

\title{Barcodes as summary \\of loss function topology}


\author{Serguei Barannikov\Addressmark[1,3]\Emailmark[1], Alexander Korotin\Addressmark[1,2], Dmitry Oganesyan\Addressmark[1], Daniil Emtsev\Addressmark[1,4], Evgeny Burnaev\Addressmark[1,2]}

\Addresstext[1]{Skolkovo Institute of Science and Technology, Moscow, Russia}
\Addresstext[2]{Artificial Intelligence Research Institute,
  Moscow, Russia}

\Addresstext[3]{CNRS, IMJ, Paris Cité University, France}


\Addresstext[4]{ETH Zurich, Switzerland}

\Emailtext[1]{s.barannikov@skoltech.ru}

\markboth{S.Barannikov, A.Korotin, D.Oganesyan, D.Emtsev, E.Burnaev}{Barcodes as summary of loss function topology}









\maketitle

\begin{abstract}
We propose to study neural networks' loss surfaces by methods of topological data analysis.
We suggest to apply barcodes of Morse complexes to explore topology of loss surfaces. An algorithm for calculations of the loss function's barcodes of local minima is described. We have conducted experiments for calculating barcodes of local minima for benchmark functions and for loss surfaces of small neural networks. Our experiments confirm our two principal  observations for neural networks' loss surfaces. First, the barcodes of local minima are located in a small lower part of the range of values of neural networks' loss function. Secondly, increase of the neural network's depth and width  lowers the barcodes of local minima. This has some natural implications for the neural network's learning and for its generalization properties. The code can be found at \url{https://github.com/grag90/BarcodeCalc}.
\end{abstract}
\begin{altkeywords}
 loss surface, persistent homology, persistence barcodes, Morse theory, neural networks   
\end{altkeywords}




\makeatletter

\section{Introduction}

Searching for minima of the loss function is the principal strategy underlying the majority of machine learning algorithms. The graph of the loss function, which is often called  \textbf{loss surface}, typically has  complicated structure \cite{li2018visualizing,attack,phys}: non-convexity, many local minima, saddle points, flat regions. These obstacles harm the exploration of the loss surface.

The searching for minima of modern neural networks is mainly carried out by gradient descent based algorithms. How these algorithms can achieve almost zero loss despite the non-convexity of the loss function remains poorly understood. 

The global topological characteristics of the gradient flow trajectories are captured by the Morse complex via decomposing the parameter space into cells of uniform flow \cite{bott,smale,thom}.
The barcodes of Morse complex constitute the fundamental summary of the topology of the gradient vector field flow \cite{B94,viterbo:2011,viterbo:2018}.
Barcodes give a decomposition of topology change of the loss function sublevel sets into the sum of "birth"-"death" of elementary features.

We argue, see section \ref{sec-nn},  that, topologically, the  "badness" of a given local minimum, harming gradient-based learning algorithms,  can be quantified by the "lifetime" length of the minimum's segment in the barcode. For gradient-based learning algorithms, this quantity measures \textit{the obligatory penalty for moving from the given local minimum to a point with lower loss value}. 

The calculation of the barcodes for various specific functions constitutes the essence of the topological data analysis. Currently available software packages for the calculation of barcodes of functions, also called  "sublevel persistence",   are  GUDHI, Dionysus, PHAT. They are based on the general algorithm requiring construction of simplicial complex and having $O(N^3)$ worst time complexity in the number of points for computation of the lowest degree barcode. These packages can currently handle calculations of barcodes for functions defined on a grid, of up to $10^6$ points, and in dimensions up to six. Thus, all current packages experience the scalability issues.

We describe an algorithm for the computation of lowest degree barcodes for functions in arbitrary dimensions. In contrast to the mentioned grid-based methods, our algorithm works with functions defined on arbitrarily sampled \textbf{point clouds}. Point cloud based methods are known to work better than grid-based methods in optimization-related problems\cite{bergstra2012random}. To compute the lowest degree barcodes we use the fact that their definition can be reformulated in geometrical terms, see definition \nolinebreak\ref{def1} in section \ref{sec-3def}. Most currently available software packages are based on the more algebraic approach as in definition \nolinebreak\ref{defin2} from section \nolinebreak \ref{sec:2ndDef}. The principal part of our algorithm has worst time complexity $O(N\log N)$ and it was tested in dimensions up to $15$ and with the number of points of up to $10^{9}$.


The proposed methodology describes properties of the loss surface of neural networks via topological features of local minima. We emphasize that the value of the loss at a minimum is only half of its topological characteristic from the barcode. The other half can be described as the value of loss function at the 1-saddle, which is naturally associated with each local minimum, see section \ref{sec-3def}. 
The 1-saddle $q$ associated with the minimum $p$  is the point where the connected component  of the sublevel set  $\Theta_{f\leq c} = \left\{ \theta\in \Theta  \mid f(\theta)\leq c \right\}$  containing  $p$ merges  with another connected component of the sublevel set containing a \emph{lower} minimum. This correspondence  between local minima and 1-saddles, that kill a connected component of $\Theta_{f\leq c}$,  is one-to-one. 

The segment $\left[f(p),f(q)\right]$, where $q$ is the 1-saddle associated with $p$, is the invariant which the barcode  associates with the minimum $p$.  The difference $f(q)-f(p)$ is the topological invariant of the minimum, quantifying its badness for gradient-based learning algorithms. The set of all such segments for all  minima is the lowest degree  barcode of $f$.

The main contributions of the paper are as follows:

\textbf{Applying the minima barcodes and the correspondence between minima and $1-$saddles to exploration of loss surfaces.} For each local minimum $p$ there is canonically defined 1-saddle $q$ (see Section \ref{sec-3def}).   The set of all segments  $\left[f(p),f(q)\right]$, where $p$ is a local minimum and $q$ is the corresponding 1-saddle of $f$, is a robust topological invariant of loss function. It is invariant in particular under the action of homeomorphisms of $\Theta$. This set of segments is a part of the full barcode.  The full barcode gives a concise summary of the  topology of the loss function and of the global structure of its gradient flow. 

\textbf{Algorithm for calculations of the minima barcodes.} We describe and analyze an algorithm for calculation  of the barcodes of minima. The algorithm takes as an input a randomly sampled or a specifically chosen set of points and the loss function's values on this set. The algorithm then employs the HNSW procedure to calculate the graph of neighbors. The next step is the computation of the barcode of minima for the function defined on the graph. The local minima give birth to clusters of points in sublevel sets. Then algorithm works by looking at neighbors of each sampled point with lower values of the function and deciding if this point belongs to an existing cluster, gives birth to a new cluster (minimum), or unifies two or more clusters ($1$-saddle).  
The second part of the algorithm working with a function on a graph is similar to a particular $0-$dimensional case of the general algorithm described in \cite{B94}.

\textbf{Experiments confirming observations on behavior  of neural networks loss functions barcodes.}  We calculate the barcodes of minima for small fully-connected neural networks of up to three hidden layers and verify that  all segments of minima's barcode belong to a small lower part of the total range of loss function's values and that with the increase in the neural network depth the minima's barcodes descend lower.

The usefulness of our approach and algorithms is not limited to optimization problems. 
Our algorithm permits the fast computation of the persistence barcodes of many functions which were not accessible until now.
These sublevel persistence barcodes 
have been successfully applied in different disciplines: cognitive science \cite{Bubenik:2009}, cosmology \cite{sousbi:2011} to name a few, 
see e.g. \cite{reviewCh:18} and references therein.

Our framework also has applications in chemistry and material science where 1-saddle points on potential energy landscapes correspond to transition states and minima are stable states corresponding to different materials or protein foldings, see e.g. \cite{B20,Oganov:09}.

The article is structured as follows. We begin with two definitions of barcodes of local minima in section \ref{sec-3def}. Our algorithm for the calculation of barcodes is described in section \ref{sec-algorithm}. In section \ref{bench} we apply our algorithm to calculate barcodes of benchmark functions. We prove the convergence of the algorithm and demonstrate it empirically in subsection \ref{sec:converge}. In section \ref{sec-nn} we calculate barcodes of the loss functions of small neural networks and describe our principal observations.

\section{Topology of loss surfaces via barcodes}
\label{sec-3def}

Barcodes give a concise summary of topological features of functions as decomposition of change of topology of function's sublevel sets into the finite sum of ``birth''--``death'' of elementary features. We propose to apply these invariants as a tool for exploring the topology of loss surfaces. 

We describe in this section two definitions of the barcodes of minima for  piecewise-smooth continuous functions. 
In this work we concentrate on the  part of barcodes, describing the ``birth''--``death'' phenomena of connected components of the loss function's sublevel sets. The approach from this section works similarly in the context of ``almost minima'', i.e. for the critical points (manifolds) of small nonzero indexes. Such points are often the terminal points of optimization algorithms in extremely high dimensional parameter spaces of deep neural networks \cite{attack}.

\textbf{First definition: merging with  connected component of a lower minimum.}

Let $f$ be a piecewise-smooth continuous function.
The values of parameter $c$ at which the topology of sublevel sets  
\begin{equation*}
\Theta_{f\leq c}= \left\{ \theta\in \Theta  \mid f(\theta)\leq c \right\} \label{thetac}    
\end{equation*}
changes are critical values of $f$. 

\begin{figure}[!htb]
\centering
\subfloat[]{\includegraphics[width=0.32\linewidth]{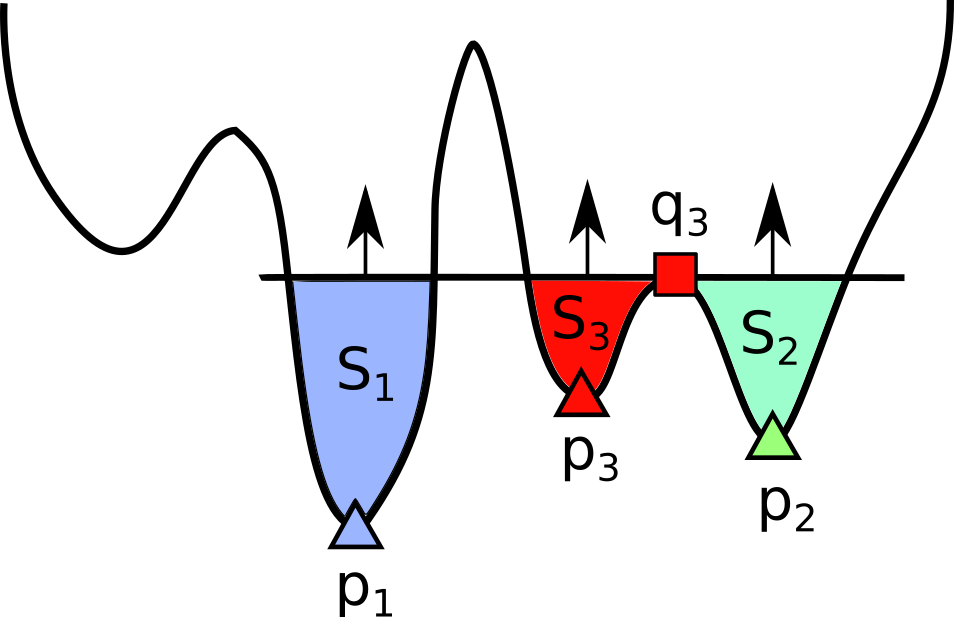}}\hfil
\subfloat[
\label{fig:1d-water-2}]{\includegraphics[width=0.32\linewidth]{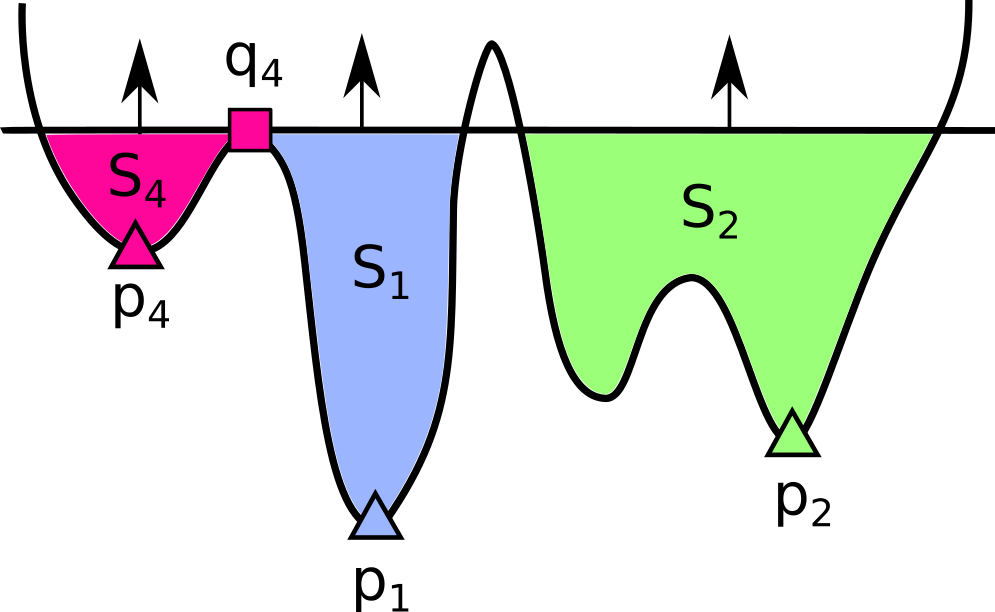}}\hfil 
\subfloat[
\label{fig:1d-water-3}]{\includegraphics[width=0.32\linewidth]{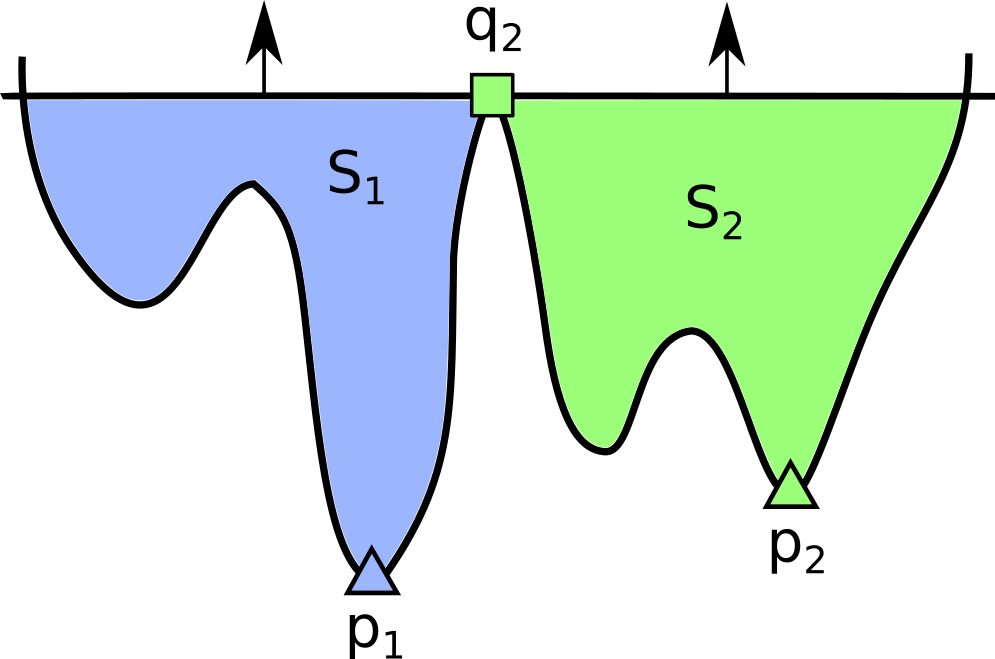}} 
\caption{Merging of connected components of sublevel sets at 1-saddles. (a)"Death" of the connected component $S_3$, the connected component $S_3$ of sublevel set merges with connected component $S_2$ at 1-saddle $q_3$, 
1-saddle $q_3$ is associated with the minimum $p_3$. (b)"Death" of the connected component $S_4$, the connected component $S_4$ of sublevel set merges with connected component $S_1$ at 1-saddle $q_4$, 1-saddle $q_4$ is associated with the minimum $p_4$. (c)"Death" of the connected component $S_2$, the connected component $S_2$ of sublevel set merges with connected component $S_1$ at 1-saddle $q_2$, 
1-saddle $q_2$ is associated with the minimum $p_2$.
 Note that the 1-saddle $q_2$ is associated with the minimum $p_2$ which is separated by another minimum from the green saddle.}
\label{fig:1d_water}
\end{figure}

Let $p$ be one of the minima of $f$. When $c$ increases from $f(p)-\epsilon$ to $f(p)+\epsilon$, a new connected component of the set $\Theta_{f\leq c}$ is born. To illustrate the process, we provide an example in Figure \ref{fig:1d_water}, where the connected components $S_1,S_2,S_3, S_4$ of  sublevel set are born at the  minima $p_1,p_2,p_3,p_4$ correspondingly.

Connected components of sublevel sets merge at 1-saddle critical points. 

A point $q$ is a \emph{1-saddle} critical point
if the intersection of the set  $\Theta_{f<f(q)}$ with any small neighborhood of $q$ has more than one connected component. 
Without loss of generality, we may assume in this section that the $1-$saddle points of $f$ are generic so that these intersections have no more than two connected components. Otherwise, one can always add a small perturbation to $f$ or consider the $1-$saddles with multiplicity given by the number of these components minus one. 

Let $p$ be a minimum, which is not global.  When $c$ is increased sufficiently, the connected component of $\Theta_{f\leq c}$ born at $p$ merges with some other connected component. Then this unified connected component may merge again with another one. After each merging, the minimum of the restriction of $f$ to the unified connected component is the smallest of the two minima of restriction of $f$ to each of the two connected components before merging. In other words, the connected component with lower minimum "swallows" at the merging point the connected component with the higher minimum, see fig \ref{fig:1d_water}.
Let $q$ be the merging point where the connected component with minimum $p$ is swallowed by a connected component whose minimum is \emph{lower}. Note that the intersection of the set  $\Theta_{f<f(q)}$ with any small neighborhood of $q$ has at least two connected components. 
\begin{definition}\label{def1}
The merging point $q$, where the connected component with minimum $p$ is swallowed by connected component with a lower minimum,
is the 1-saddle associated naturally with the minimum $p$. The segment  $\left[f(p),f(q)\right]$ is the invariant associated in barcode with minimum $p$.
\end{definition}
Note that the two connected components of the intersection of a small neighborhood of such $q$ with $\Theta_{f<f(q)}$ belong to two different connected components of the whole set $\Theta_{f< f(q)}$. The 1-saddles of this type are called ``+'' (``plus'') or ``death'' type.  

\begin{proposition}
The described correspondence between local minima and 1-saddles of this type is one-to-one. 
\end{proposition}
\begin{proof}
The correspondence in the opposite direction can be described as follows.
Let $q$ be a 1-saddle point of such type that the two branches of the set $\Theta_{f<f(q)}$ near $q$ are not connected in the whole set
$\Theta_{f<f(q)}$. One of the connected components of the sublevel set $\Theta_{f\leq c}$ splits into two when $c$ decreases from $f(q)+\epsilon$ to $f(q)-\epsilon$.  Let $p_1$ and $p_2$ be the two minima of the restriction of $f$ to each of these two  connected components. Let $p_1>p_2$
be the highest of the two minima.
The 1-saddle $q$ is associated by the Definition 1 with the local minimum $p_1$, since the two connected components merge at $q$.
\begin{equation*}
    p_1\not\in \bigg\{\operatorname{arg} \min_{\substack{x\in \text{connected component}\\ \text{of }   \Theta_{f\leq f(q)+\epsilon} }} f(x) \bigg\}
\end{equation*}
Notice that $p_1$ did not appear as a minimum of $f$ on one of connected components of $\Theta_{f\leq f(q)+ \epsilon}$. In other words the minimum $p_1$, that corresponds to $q$ by definition 1, is the new minimum  appearing in the set of minima of $f$ on connected components of $\Theta_{f\leq c}$ when $c$ decreases from $f(q)+ \epsilon$ to $f(q)-\epsilon$. 
\end{proof}
It is important for application in section \ref{subs:Dnn} that 1-saddles associated with minima
are described in the following way.
\begin{proposition} \label{prop:Paths}
Consider various paths $\gamma$ starting from local minimum $p$ and going to a lower minimum. Let $m_{\gamma}\in\Theta$ be the maximum of the restriction of $f$ to such path $\gamma$. Then    
 $1$-saddle $q$ corresponding to the local minimum $p$ in the barcode is the minimum over the set of all such paths $\gamma$ of the maxima $m_{\gamma}$:
 \begin{equation}q=\operatorname{arg}\bigg[ \min_{\substack{\gamma: [0,1]\rightarrow \Theta \\ \gamma(0)=p,\,f(\gamma(1))<f(p)}}\max_t f\big(\gamma(t)\big)\bigg] \label{eq:Paths}
 \end{equation}
\end{proposition}
\begin{proof}
Let the connected component of local minimum $p$ merge with the connected component of a lower minimum at the point $q$. Then there is a path in the sublevel set $\Theta_{f\leq f(q)}$  passing through $q$ and connecting $p$ with the lower minimum. The point $q$ is the maximum of the restriction of $f$ to such a path.  And there is no path in sublevel sets $\Theta_{f\leq c}$, $c<f(q)$ , that connects $p$ with a lower minimum. Therefore there is no path with $f(m_\gamma)<f(q)$ and $q$ satisfies equation (\ref{eq:Paths}).
\end{proof}

\textbf{Second definition: invariants of filtered complexes.}\label{sec:2ndDef}

We recall here the definition for the full barcode \cite{B94,viterbo:2011}. Although our algorithm from section \ref{sec-algorithm} is based on definition \ref{def1}, we need definition \ref{defin2} below to put things into the general framework in which these invariants constitute the full description of the loss function topology and of the global behaviour of the gradient flow. 

Families of gradient flow trajectories emanating from the same singular point decompose the domain $\Theta$ into cells on which the gradient flow behaves uniformly\cite{thom}.

Chain complex is the algebraic counterpart of an intuitive idea representing complicated geometric objects as a decomposition into simple pieces. It converts such a decomposition into a collection of vector spaces and linear maps.

Recall that a chain complex $(C_{*},\partial_{*})$ is a sequence
of finite-dimensional vector spaces $C_j$ (spaces of "$j-$chains") and linear operators ("differentials")
\[
\rightarrow C_{j+1}\stackrel{\partial_{j+1}}{\rightarrow}C_{j}\stackrel{\partial_{j}}{\rightarrow}C_{j-1}\rightarrow\ldots\rightarrow C_{0},
\]
which satisfy $\partial_{j}\circ\partial_{j+1}=0$. 
The $j-$th homology of the chain complex $(C_{*},\partial_{*})$
is the quotient of vector spaces
$H_{j}=\ker\left(\partial_{j}\right)/\textrm{im}\left(\partial_{j+1}\right)$.

The decomposition of complicated geometric object into simple pieces is often done  in certain consecutive order,  in that case its algebraic counterpart is  $\mathbb{R}-$filtered  chain complex. A subcomplex $(C'_{*},\partial'_{*})\subseteq (C_{*},\partial_{*})$ is a sequence of subspaces  $C'_{j}\subseteq C_{j}$ equipped with compatible differentials $\partial'_{j}= \partial_{j} \mid_{C'_{j}}$. A  chain complex $C_{*}$  is called 
 $\mathbb{R}-$filtered  if  $C_{*}$  is equipped with an increasing
 sequence of subcomplexes ($\mathbb{R}-$filtration): $F_{s_1}C_{\ast}\subset F_{s_2}C_{*}\subset\ldots  \subset F_{s_{max}}C_{*}=C_{*}$, indexed by a finite set of real numbers $s_1<s_2<\ldots<s_{\text{max}}$. 
 \begin{theorem}\label{theorem1} (\cite{B94}, Section 2)
Any $\mathbb{R}-$filtered chain complex  $C_{*}$
 can be brought to canonical form, a canonically defined direct
sum of $\mathbb{R}-$filtered complexes of two types: one-dimensional
complexes with trivial differential $\partial_{j}(e_{i})=0$ and two-dimensional
complexes with trivial homology $\partial_{j}(e_{i_2})=e_{i_1}$, by a linear transformation preserving
the $\mathbb{R}-$filtration. The resulting direct sum of such simple $\mathbb{R}-$filtered complexes  is unique.
\end{theorem}

\begin{definition}\label{defin2}
The full barcode is a visualization of the  decomposition of an $\mathbb{R}-$filtered complex according to the theorem \ref{theorem1}.
Each filtered 2-dimensional complex with trivial homology $\partial_{j}(e_{i_2})=e_{i_1}$, $\left\langle {e}_{i_{1}}\right\rangle =F_{\leq s_{1}}$,$\left\langle {e}_{i_{1}},{e}_{i_{2}}\right\rangle =F_{\leq s_{2}}$ 
describes a single topological feature in dimension $j$ which is  "born" at  $s_{1}$ and which "dies" at $s_{2}$. It is represented by  segment $[s_1,s_2]$ in the degree-$j$ barcode. And each filtered one-dimensional complex with trivial differential,
$\partial_j{e}_{i}=0$ , $\left\langle {e}_{i}\right\rangle =F_{\leq r}$ describes a topological feature in dimension $j$ which is "born" at $r$ and never "dies". It is represented by the half-line $[r,+\infty[$ in the degree-$j$ barcode.
\end{definition}

The proof of Theorem \ref{theorem1} is given  in \cite{B94}, Section 2, see also \cite{viterbo:2011,viterbo:2018}. We describe it in Appendix \ref{proofTh1} for the reader's convenience.  Essentially, one can bring an $\mathbb{R}-$filtered complex to the required direct sum of simple $\mathbb{R}-$filtered complexes  by induction, starting from the lowest basis elements of degree
one, in such a way that the manipulation of degree $j$ basis elements
does not destroy the obtained decomposition into simple $\mathbb{R}-$filtered complexes in degree $j-1$ and in lower filtration pieces in degree $j$. 

Let $f:\Theta\to \mathbb{R}$ be a smooth, or more generally, piece-wise smooth continuous function  such that the sublevel sets $\Theta_{f\leq c} = \left\{ \theta\in \Theta  \mid f(\theta)\leq c \right\}$ are compact. 

There are different filtered chain complexes computing the homology of the topological spaces $\Theta_{f\leq c}$,  the \v{C}ech complexes, the simplicial complexes, or the  CW-complexes. Without loss of generality, the piece-wise smooth function $f$ can be assumed smooth, otherwise one can always replace $f$ by a smooth approximation. By adding a small perturbation we can also assume that critical points of $f$ are non-degenerate.

One filtered chain complex naturally associated with such function $f$, with subcomplexes $F_{s}C_{\ast}$  computing  homology of sublevel sets $\Theta_{f\leq s}$,  is the Morse complex, see appendix \ref{subsMorse} and \cite{bott,smale} for more details.   
The Morse complex is defined as follows. The basis elements in 
the $k-$vector spaces $C_j$ are in one-to-one correspondence with
the critical points of $f$ of index $j$ equipped with a choice of orientation. The orientation here is a choice of the orientation of $j$-dimensional subspace of tangent space at the critical point, on which the Hessian is negative-definite. The matrix of $\partial_{j}$ consists of the numbers of gradient trajectories, counted with signs, between the index $j$ and the index $(j-1)$ critical points. The natural filtration on Morse complex is  given by the values of critical points. Basis elements in $F_{s}C_{\ast}$ correspond to critical points with critical value $s$ and lower. 

\begin{proposition}
Let $p$ be a minimum, which is not global.
Then $p$ represents trivial homology class in Morse complex computing the homology of $\Theta_{f\leq c}$ for big enough $c$ , i.e. $p$ is a lower basis element of one of the two-dimensional complexes with trivial homology in the canonical form. Thus $p$  is coupled with a 1-index saddle $q$. 
This is the 1-saddle from definition \nolinebreak\ref{def1}, i.e. $q$ is the 1-saddle at which the sublevel set connected component corresponding to $p$ is swallowed by a connected component with lower minimum. The segment $\left[f(p),f(q)\right]$  then corresponds to the minimum $p$ by definition \nolinebreak\ref{def1}.
\end{proposition}
\begin{proof} Let $q$ be  the 1-saddle at which the connected component corresponding to $p$ is swallowed by a connected component with a lower minimum $r$.  The  homology 
$H_0 (\Theta_{f\leq f(q)-\epsilon})$ are generated linearly by classes of connected components of  $\Theta_{f\leq f(q)-\epsilon}$. They  correspond to the generators of the Morse complex given by minima of restriction of $f$ to each connected component.  In the Morse complex computing homology of  $\Theta_{f\leq f(q)+\epsilon}$ ,  the generator corresponding to the local minimum $p$ equals to  the boundary of the generator corresponding to the  1-saddle $q$ plus the generator corresponding to the lower minimum $r$ plus perhaps boundaries of  lower than $q$ saddles.  Therefore  $q$ is coupled with $p$ in the canonical form.
\end{proof}

Similar results hold for other types of filtered complexes representing sublevel homology, like cubical or CW-complexes, for continuous piece-wise smooth functions (see e.g. \cite{chazal2011scalar, cohen2007stability}). 


The total number of different topological features in sublevel sets $\Theta_{f\leq c}$ of the loss function can be read immediately from the barcode. Namely, the number of intersections of  horizontal line at level $c$ with segments in the index $j$  barcode  gives  the number of independent topological features of dimension $j$ in $\Theta_{f\leq c}$, see examples in section \ref{bench}.

\section{An Algorithm for Calculation of Minima Barcodes}
\label{sec-algorithm}
In this section, we describe an algorithm for the calculation of the barcodes of local minima. The algorithm uses definition \nolinebreak\ref{def1} of barcodes from Section \ref{sec-3def} that is based on the evolution on the connected components of the sublevel sets.

To analyse the surface of the given function $f:\Theta\rightarrow \mathbb{R}$, we first build its approximation by finite graph-based construction. To do this, we consider a randomly sampled subset of points $\{\theta_1, \ldots, \theta_N\}\in \Theta$ and construct a graph with these points as vertices.  We connect vertices with an edge if the points are close. Thus, for every vertex $\theta_{n}$, by comparing $f(\theta_{n})$ with $f(\theta_{n'})$ for neighbors $\theta_{n'}$ of $\theta_{n}$, we are able to understand the local topology near the point $\theta_{n}$. At the same time,  connected components of sublevel sets $\Theta_{f\leq c}$
correspond, see  proposition \ref{prop:convrg} below, to connected components of the subgraph $\Xi_{f\leq c}$ of points $\theta_{n}$, such that $f(\theta_{n})\leq c$.

Two technical details here are the choice of points $\theta_{n}$ and the definition of closeness, i.e. when to connect points by an edge. In our experiments, we sample points uniformly from some rectangular box of interest. To add edges, we compute the oriented $k$-nearest neighbor graph on the given points,  then drop the orientation of edges and check that the distance between  neighbors does not exceed $c(D)N^{-\frac{1}{D}}$, where $D$ is the dimension of $f$'s input. We use $k=2D$ in our experiments




We describe now the part of the algorithm that computes barcodes of a function from its  graph-based approximation described above. The key idea is to monitor the evolution of the connected components of the sublevel sets of the graph $\Xi_{f\leq c}=\{\theta_{n}\text{ | }f(\theta_{n})\leq c\}$ for increasing $c$. 

\begin{algorithm}[!htb]
\SetAlgorithmName{Algorithm}{empty}{Empty}
\SetKwInOut{Input}{Input}
\SetKwInOut{Output}{Output}
\Input{ Undirected graph $G=(V,E)$; function $f$ on graph vertices.}
\Output{ Barcodes: a list of "birth"-"death" pairs.}
$S\leftarrow \{\}$\;
\For{$\theta\in V\text{ in increasing order of }f(\theta)$}{
	$S'\leftarrow \{s\in S\text{ | }\exists \theta'\in s\text{ such that }(\theta,\theta')\in E\text{ and }f(\theta)>f(\theta')\}$\;
	\uIf{$S'=\emptyset$}{
      	$S\leftarrow S\sqcup \{\{\theta\}\}$\;
    }
    \Else{
        $f^{*}\leftarrow\min f(\theta')$ for $\theta'\in \bigsqcup\limits_{s\in S'} s$\;
        \For{$s\in S'$}{
            $f^{s}\leftarrow \min f(\theta')$ for $\theta'\in s$\;
            \uIf{$f^{s}\neq f^{*}$}{
              	$\text{Barcodes}\leftarrow \text{Barcodes}\sqcup \{\big(f^{s}, f(\theta)\big)\}$\;
            }
        }
        $s_{\text{new}}\leftarrow \big(\bigsqcup\limits_{s\in S'}s\big) \sqcup \{\theta\}$\;
        $S\leftarrow (S\setminus S')\sqcup \{s_{\text{new}}\}$\;
    }
 }
 \For{$s\in S$}{$f^{s}\leftarrow \min f(\theta')$ for $\theta'\in s$\;
 $\text{Barcodes}\leftarrow \text{Barcodes}\sqcup \{\big(f^{s}, \infty \big)\}$\;
 }
\textbf{return} Barcodes
\caption{Barcodes of minima computation for function on a graph.}
\label{algorithm-graph}
\end{algorithm}

For simplicity we assume that points $\theta$ are ordered w.r.t. the value of function $f$, i.e. for $n<n'$ we have $f(\theta_{n})<f(\theta_{n'})$. In this case we are interested in the evolution of connected components throughout the process of sequential adding of vertices $\theta_{1},\theta_{2},\dots, \theta_{N}$ to graph, starting from an empty graph. We denote the subgraph on vertices $\theta_{1},\dots,\theta_{n}$ by
$\Xi_{n}$. When we add new vertex $\theta_{n+1}$ to $\Xi_{n}$, there are three possibilities for connected components to evolve:
\begin{itemize}
    \item[1.] Vertex $\theta_{n+1}$ has zero degree in $\Xi_{n+1}$. This means that $\theta_{n+1}$ is a local minimum of $f$ and it forms a new connected component in the sublevel set.
    \item[2.] All the neighbors of $\theta_{n+1}$ in $\Xi_{n+1}$ belong to one connected component in $\Xi_{n}$.
    \item[3.] All the neighbors of $\theta_{n+1}$ in $\Xi_{n+1}$ belong to $K\geq 2$ connected components $s_{1},s_{2},\dots, s_{K}$ in $\Xi_{n}$. Thus, all these components will form a single connected component in $\Xi_{n+1}$.
\end{itemize}
In the third case, according to definition \nolinebreak\ref{def1} of section \ref{sec-3def}, the point $\theta_{n+1}$ is  a discrete $1$-saddle point. Thus, one of the components $s_{k}$ swallows all the rest. This is the component which has the lowest minimal value. For other components this gives their barcodes: for $s_{i}, i\neq k$ the birth-death pair is $\big[\min\limits_{\theta\in s_{i}}f(\theta); f(\theta_{n+1})\big]$.
We summarize the procedure in the algorithm \ref{algorithm-graph}.

In the practical implementation of the algorithm, we precompute the values of function $f$ at the vertices of $G$. Besides that, we use the disjoint set data structure to store and join connected components during the process. We also keep and update the global minima in each component. We did not include these tricks into the algorithm's pseudo-code in order to keep it simple.

\begin{remark}
Given a function on a graph one can construct an $\mathbb{R}-$filtered chain complex $C_1\overset{\partial_1}{\to} C_0$ as follows. The basis of the vector space of $0-$chains $C_0$ is given by the set of graph's vertices and the basis of the space of $1-$chains $C_1$ is given by the set of graph's edges, for simplicity we consider the vector spaces over the field $\mathtt{k}=\{0,1\}$. Then the differential of an edge is the sum of its two ends. The filtration is defined by the values of the function. The filtration for an edge is given by the maximum of the function's values on the two  ends of the edge. For this particular chain complex the general algorithm from\cite{B94}, that calculates the barcodes of $\mathbb{R}-$filtered complexes, simplifies and its $0-$dimensional part essentially gives the described algorithm for barcodes of minima for function on a graph.
\end{remark}

The resulting complexity of the algorithm's principal part is $O(N\log N)$ in the number of points. Here it is important to note that the procedure of graph creation may be itself time-consuming. In our case, the most time consuming operation is nearest neighbor search. In our code, we used efficient HNSW Algorithm for approximate NN search by \cite{malkov2018efficient}.
Also since we only take neighbors lying no further than  fixed small distance $r=O(N^{-\frac{1}{D}}) $,  one can use the following simple strategy as well. First, distribute  points from the sample over boxes of fixed grid with edges of size $r$. Then, in order to determine the neighbors, check distances from each point to the points lying only in the same box and in the neighboring boxes of the grid.

\section{Barcodes of benchmark functions and convergence of the algorithm}
\label{benc-conv}

In this section we apply our algorithm to describing the topology of test functions. In subsection \ref{bench} we apply the algorithm to visual examples and
in subsection \ref{sec:converge} we
check the convergence on benchmark functions. In section \ref{sec-nn} we apply our algorithm to analysis of loss surfaces of \textbf{small neural networks}.

We apply our algorithm to several functions ${f:\mathbb{R}^{D}\rightarrow \mathbb{R}}$ from \textbf{Global Optimization Benchmark} (see e.g. \cite{jamil2013literature}). These functions are designed to fool global optimization algorithms,  they are very complex, have many local minima and saddle points even for  small dimensions. Thus, the computation of barcodes and minimum-saddle correspondence for these functions is also challenging. 

\textbf{Barcodes of 2D benchmark functions} \label{bench}

\begin{figure}[h]
\centering
\subfloat[Surface plot (3D).\label{fig:humpcamel6plot3d}]{\includegraphics[width=0.33\linewidth]{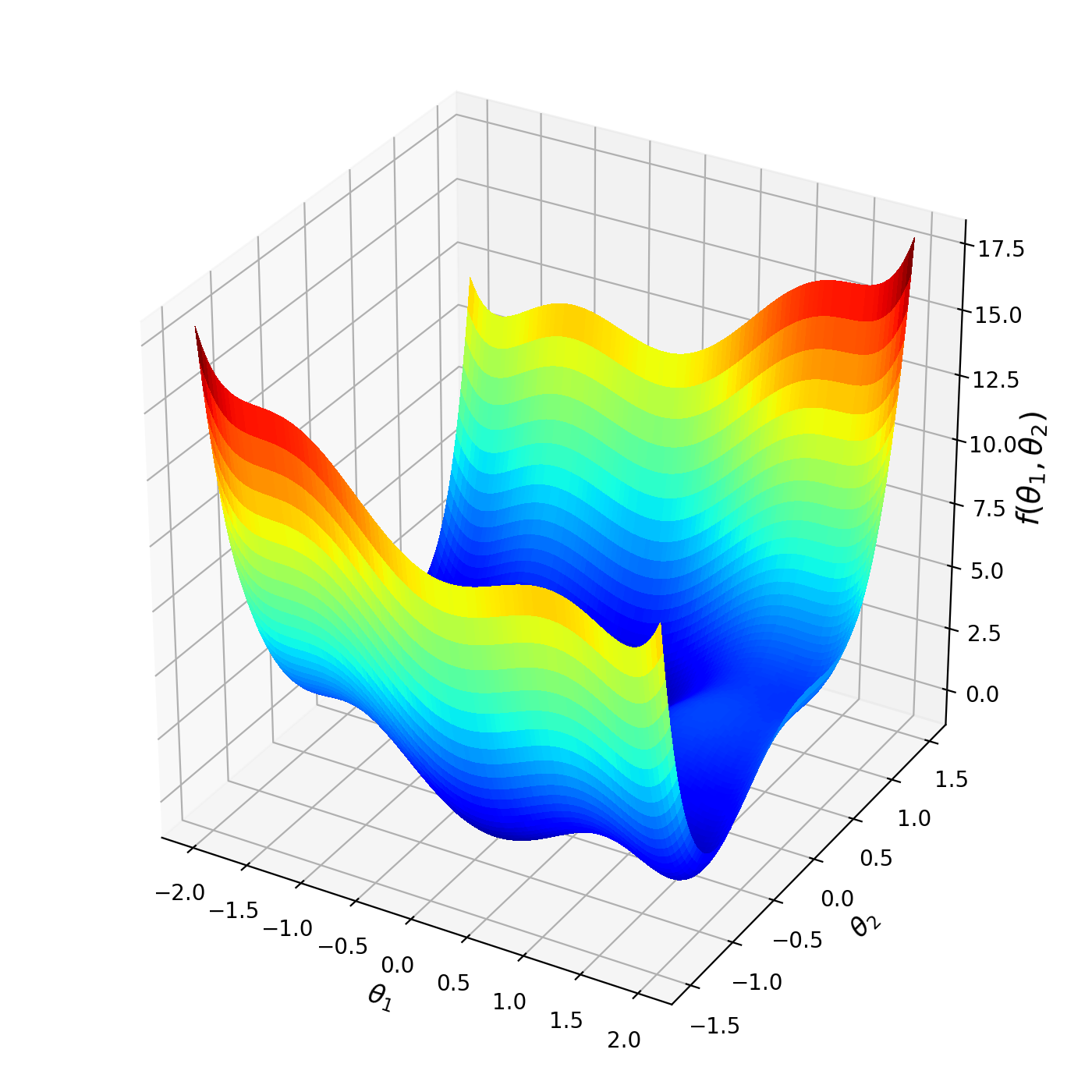}}\hfil
\subfloat[Correspondence between minima and 1-saddles \label{fig:humpcamel6plot2d}]{\includegraphics[width=0.33\linewidth]{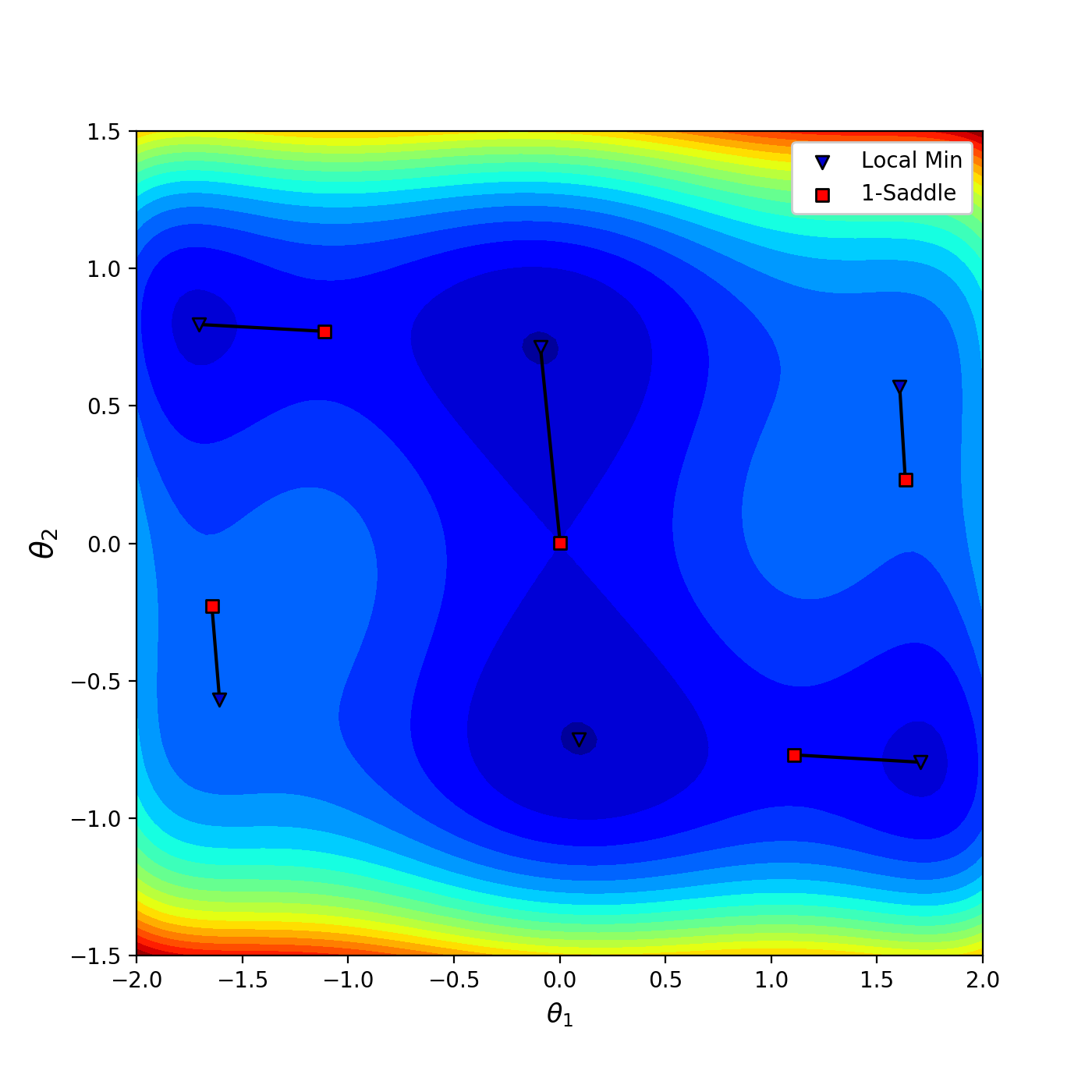}}\hfil 
\subfloat[Barcode of local minima.\label{fig:humpcamel6barcodes}]{\includegraphics[width=0.33\linewidth]{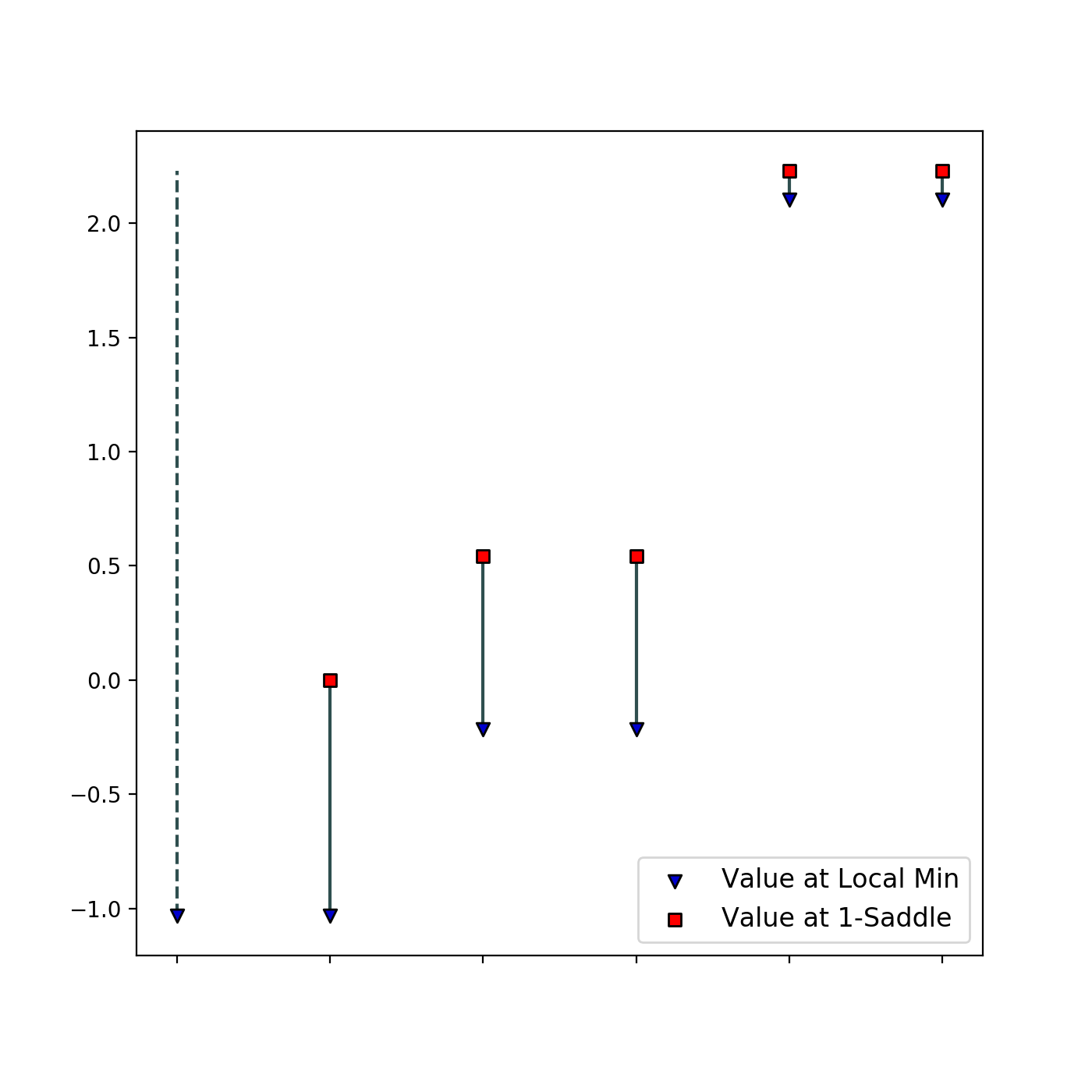}} 

\caption{HumpCamel6 function, its minima-saddle correspondence and barcode computed by Algorithm \ref{algorithm-graph}.}
\label{fig:humpcamel6plot}
\end{figure}

To begin with, we compute barcodes of several $2$-dimensional benchmark functions. We visualise obtained barcodes and minima-saddles correspondence. Next, we conduct the experiments to estimate how the number of points used to compute barcodes influences the quality of the answer. We consider the following test objective functions:
\begin{enumerate}
    \item \textbf{HumpCamel6 } function $f:[-2,2]\times [-1.5, 1.5]\rightarrow \mathbb{R}$ with 6 local minima (Figure \ref{fig:humpcamel6plot}):
    \begin{equation*}
        f(\theta_1, \theta_2) = (4 - 2.1\theta_1^2 + \theta_1^4/3)\theta_1^2 + \theta_1\theta_2 + (-4 + 4\theta_2^2)\theta_2^2
         \label{func-humpcamel6}
    \end{equation*}
\end{enumerate}

\begin{figure}[h!]
\centering
\subfloat[Surface plot (3D).\label{fig:langermann6plot3d}]{\includegraphics[width=0.33\linewidth]{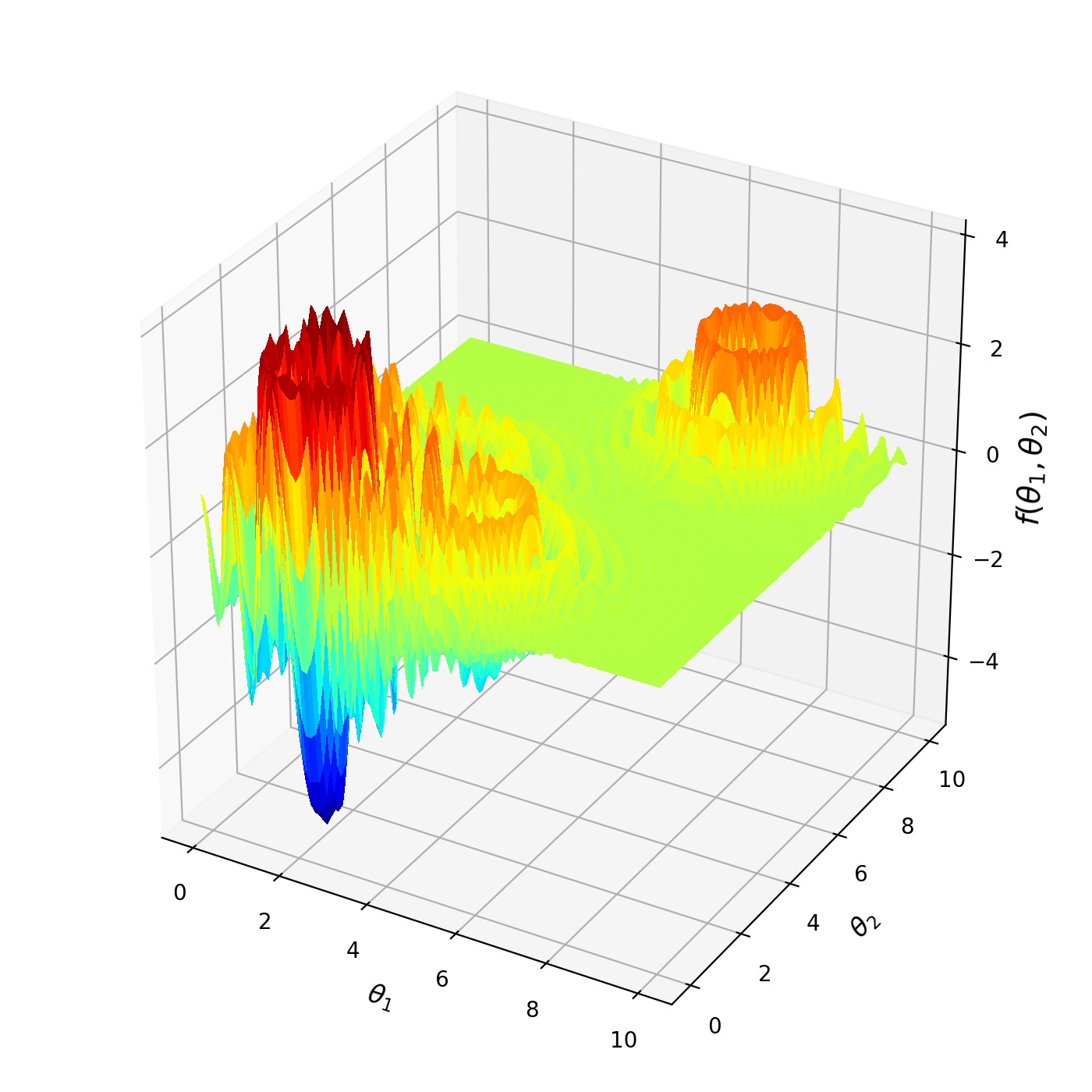}}\hfil
\subfloat[Correspondence between minima and 1-saddles \label{fig:langermann6plot2d}]{\includegraphics[width=0.33\linewidth]{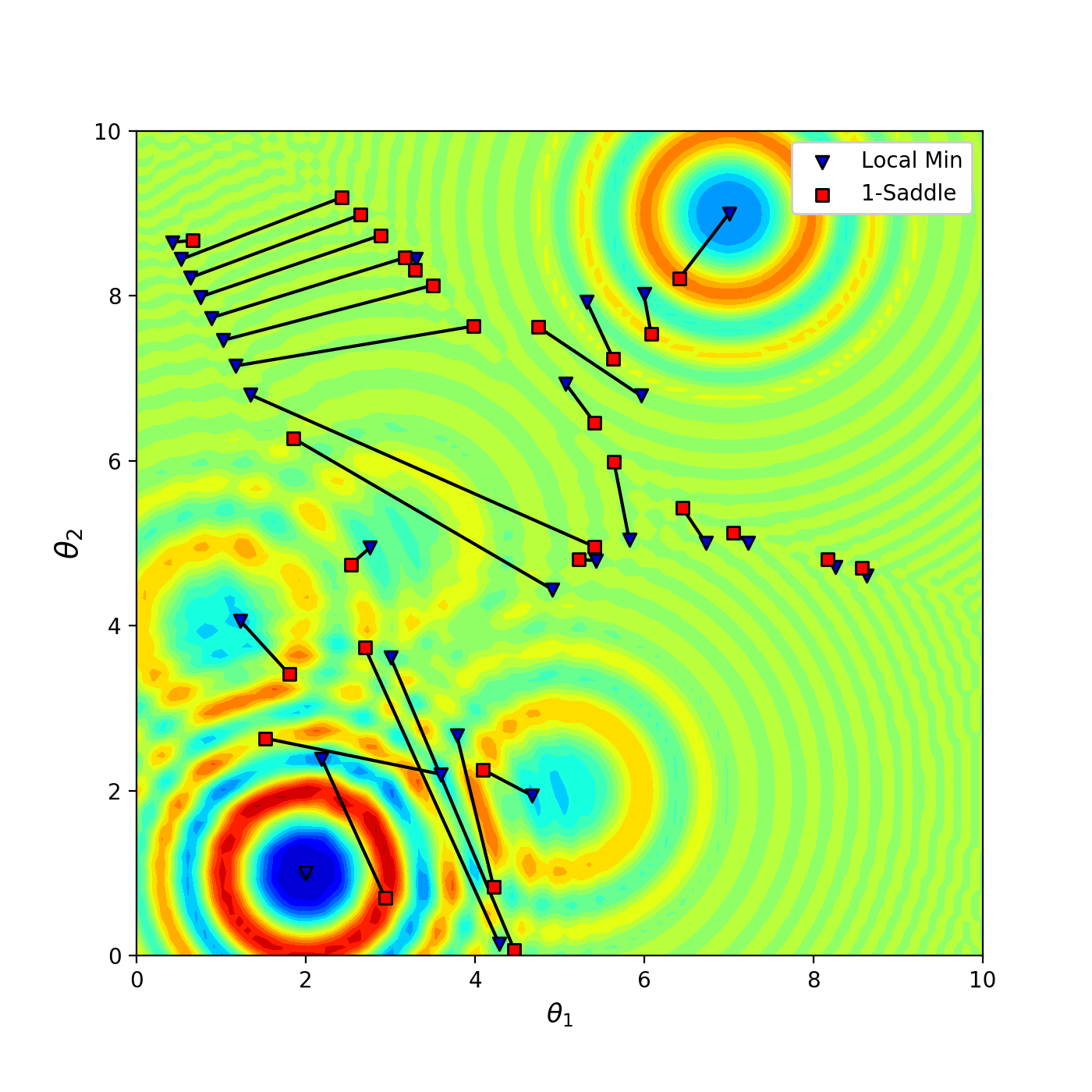}}\hfil 
\subfloat[Barcode of local minima.\label{fig:langermann6barcodes}]{\includegraphics[width=0.33\linewidth]{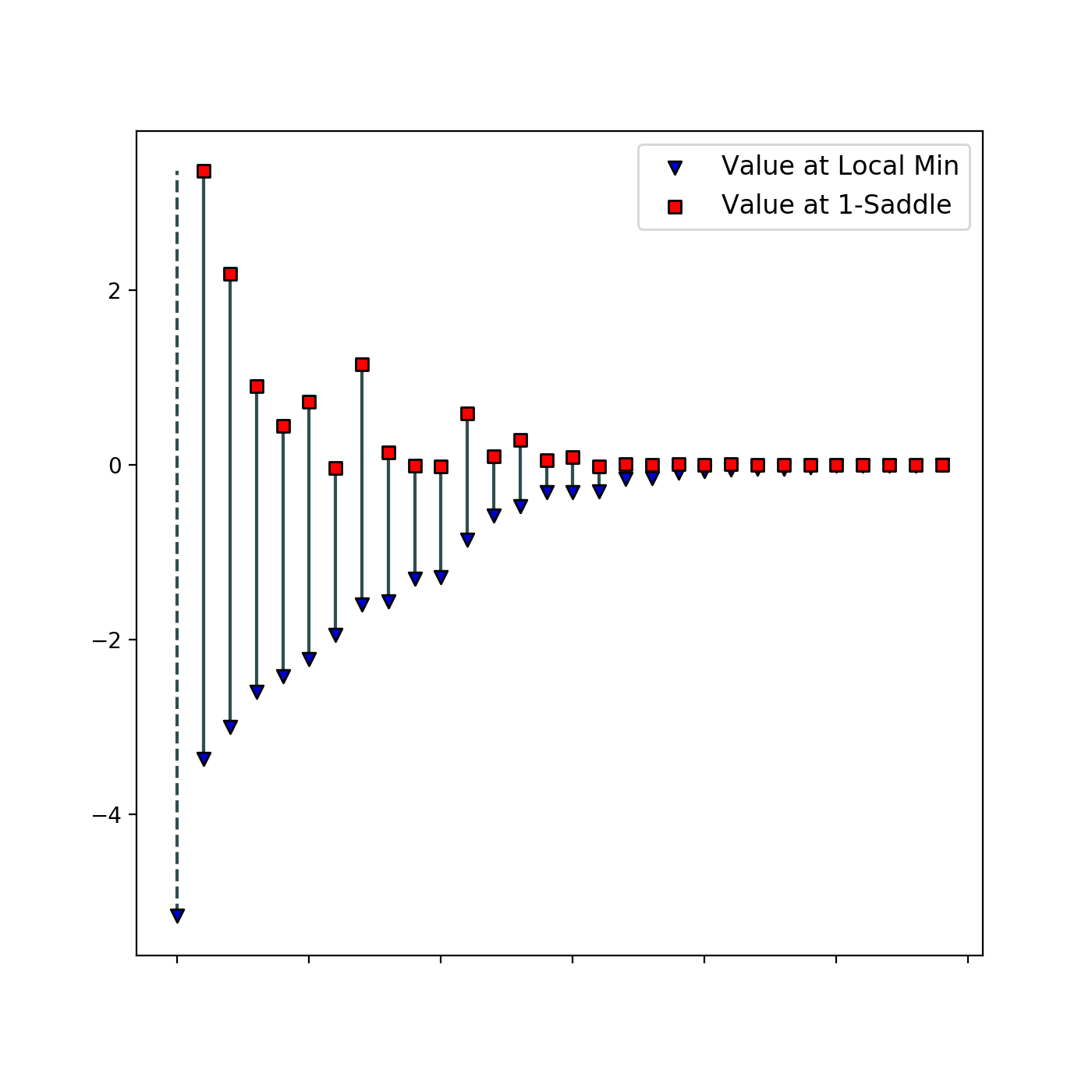}} 
\caption{Langermann function, its minima-saddle correspondence and barcodes computed by Algorithm \ref{algorithm-graph}.}
\label{fig:langermann-plot}
\end{figure}

\begin{enumerate}
  \setcounter{enumi}{1}
    \item \textbf{Langermann} test objective function $f:[0,10]^{2}\rightarrow \mathbb{R}$ (Figure \ref{fig:langermann-plot}):
    \begin{equation*}
        f(\theta_{1},\theta_{2})=-\sum_{i=1}^{5}\frac{c_{i}\cos\big(\pi \big[(\theta_{1}-a_{i})^2+(\theta_{2}-b_{i})^{2}\big]\big)}{\exp{(\frac{1}{\pi} \big[ (\theta_{1}-a_{i})^{2}+(\theta_{2}-b_{i})^{2}\big] )}}.
        \label{func-langermann}
    \end{equation*}


    To keep the plots simple, we displayed minimum-saddle correspondence only for Top-30 segments in barcode sorted by the size of the cluster (number of points at the cluster death moment). As wee see in Figure \ref{fig:langermann6plot2d}, the correspondence between minimum-saddle is non-trivial. For many minima, the corresponding saddles are rather distant. In particular, this observation illustrates the remark discussed in Figure \ref{fig:1d_water}: the corresponding canonical $1$-saddle is not necessarily the nearest saddle.
\end{enumerate}

\begin{enumerate}
  \setcounter{enumi}{2}
    \item \textbf{Wavy} test objective function $f:[-\pi,\pi]^{2}\rightarrow \mathbb{R}$ (Figure \ref{fig:wavy-plot}):
    \begin{equation*}
        f(\theta_{1},\theta_{2})=1-\frac{1}{2}\sum_{d=1}^{2}\cos(10\, \theta_{d})\cdot e^{-\frac{\theta_{d}^{2}}{2}},
        \label{func-wavy}
    \end{equation*}
    \begin{figure}[!h]
\centering
\subfloat[Surface plot (3D).\label{fig:wavyplot3d}]{\includegraphics[width=0.33\linewidth]{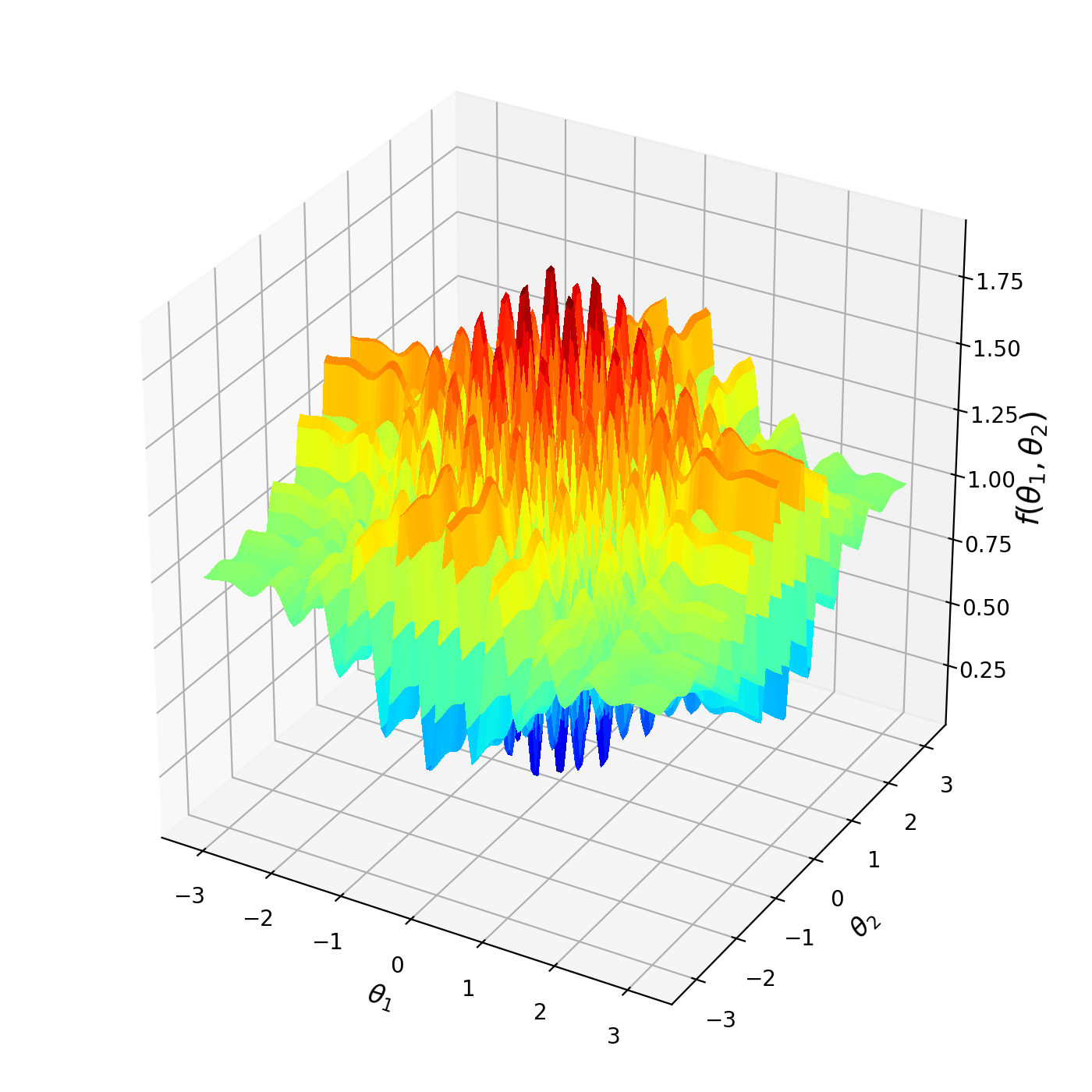}}\hfil
\subfloat[Color plot (2D) \& correspondence between minima and 1-saddle points.\label{fig:wavyplot2d}]{\includegraphics[width=0.33\linewidth]{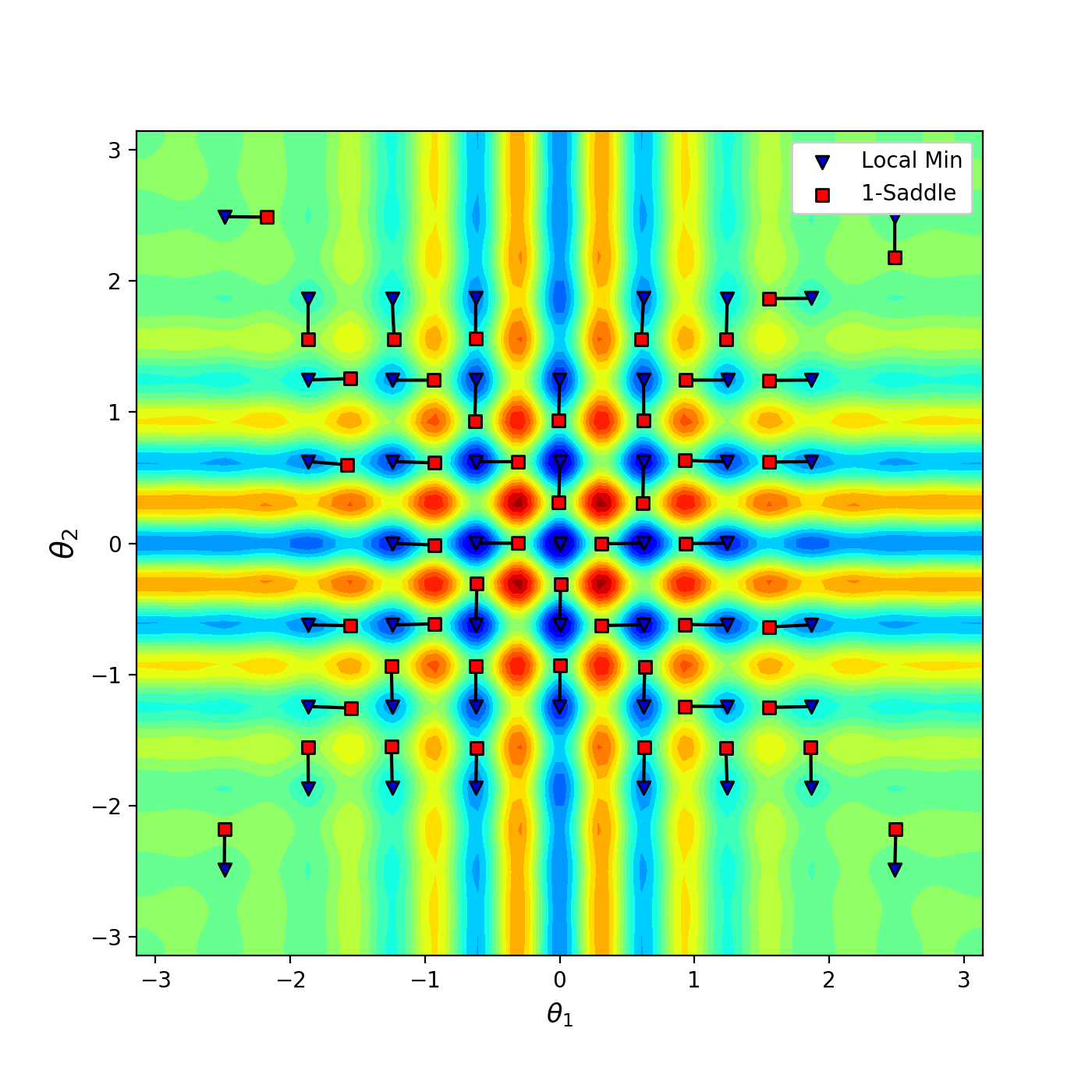}}\hfil 
\subfloat[Barcode of local minima.\label{fig:wavybarcodes}]{\includegraphics[width=0.33\linewidth]{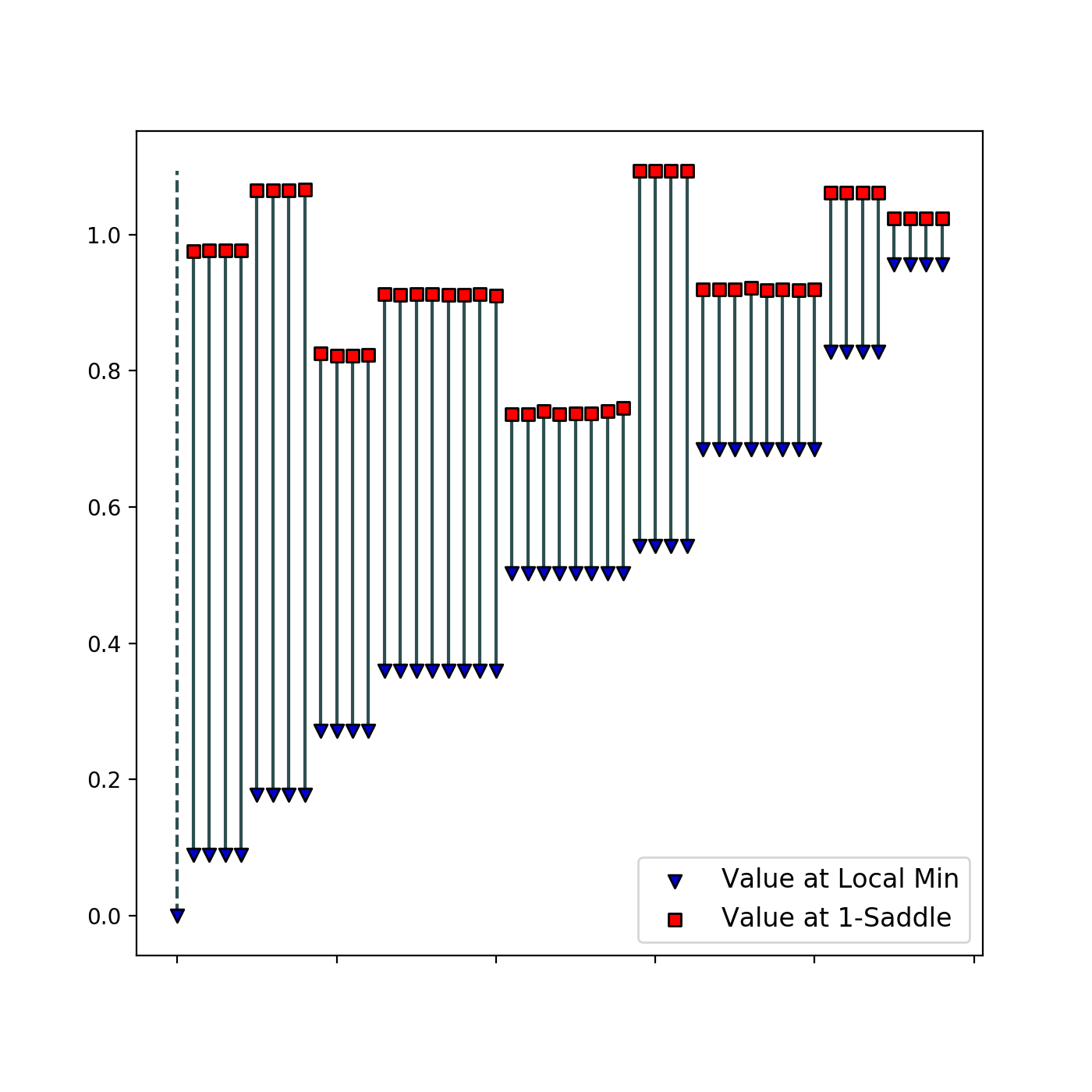}} 

\caption{Wavy function, its minima-saddles correspondence \& barcode computed by Algorithm \ref{algorithm-graph}}
\label{fig:wavy-plot}
\end{figure}
    Wavy function is symmetric with respect to the dihedral group $D_4$ of order $8$.  Thus, its minimum-saddle critical value pairs come in multiplets forming simple representations of the group $D_4$. 
\end{enumerate}

\begin{enumerate}
  \setcounter{enumi}{3}
    \item \textbf{HolderTable} test objective function $f:[-10,10]^{2}\rightarrow \mathbb{R}$ (Figure \ref{fig:holder-plot}):
    \begin{equation*}
        f(\theta_{1},\theta_{2})=-\left| e^{|1-\frac{\sqrt{\theta_{1}^{2}+\theta_{2}^{2}}}{\pi}|}\sin(\theta_{1})\cos(\theta_{2})\right|
        \label{func-holder}
    \end{equation*}
\begin{figure}[h!]
\centering
\subfloat[Surface plot (3D).\label{fig:holderplot3d}]{\includegraphics[width=0.33\linewidth]{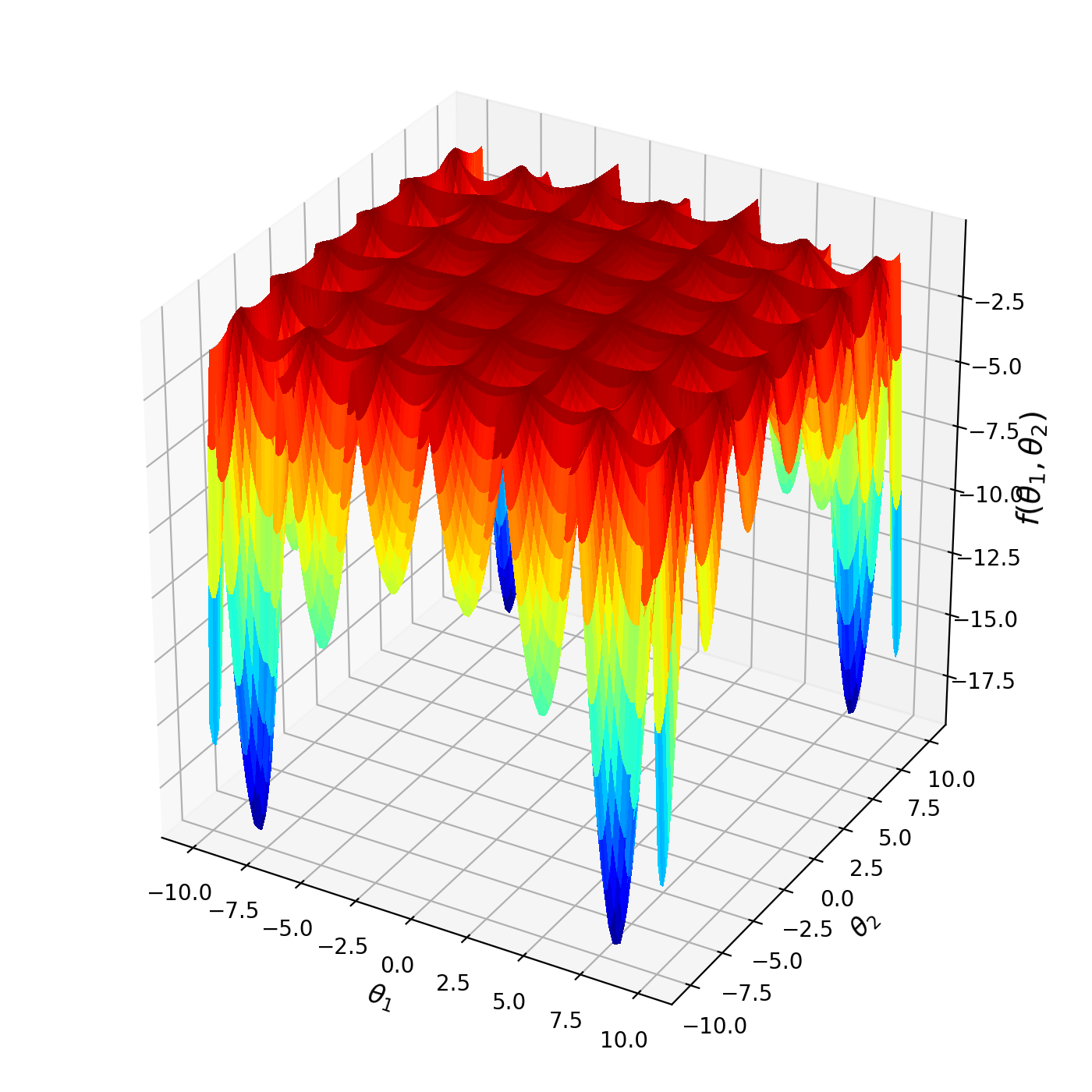}}\hfil
\subfloat[Color plot (2D) \& correspondence between minima and 1-saddle points.\label{fig:holderplot2d}]{\includegraphics[width=0.33\linewidth]{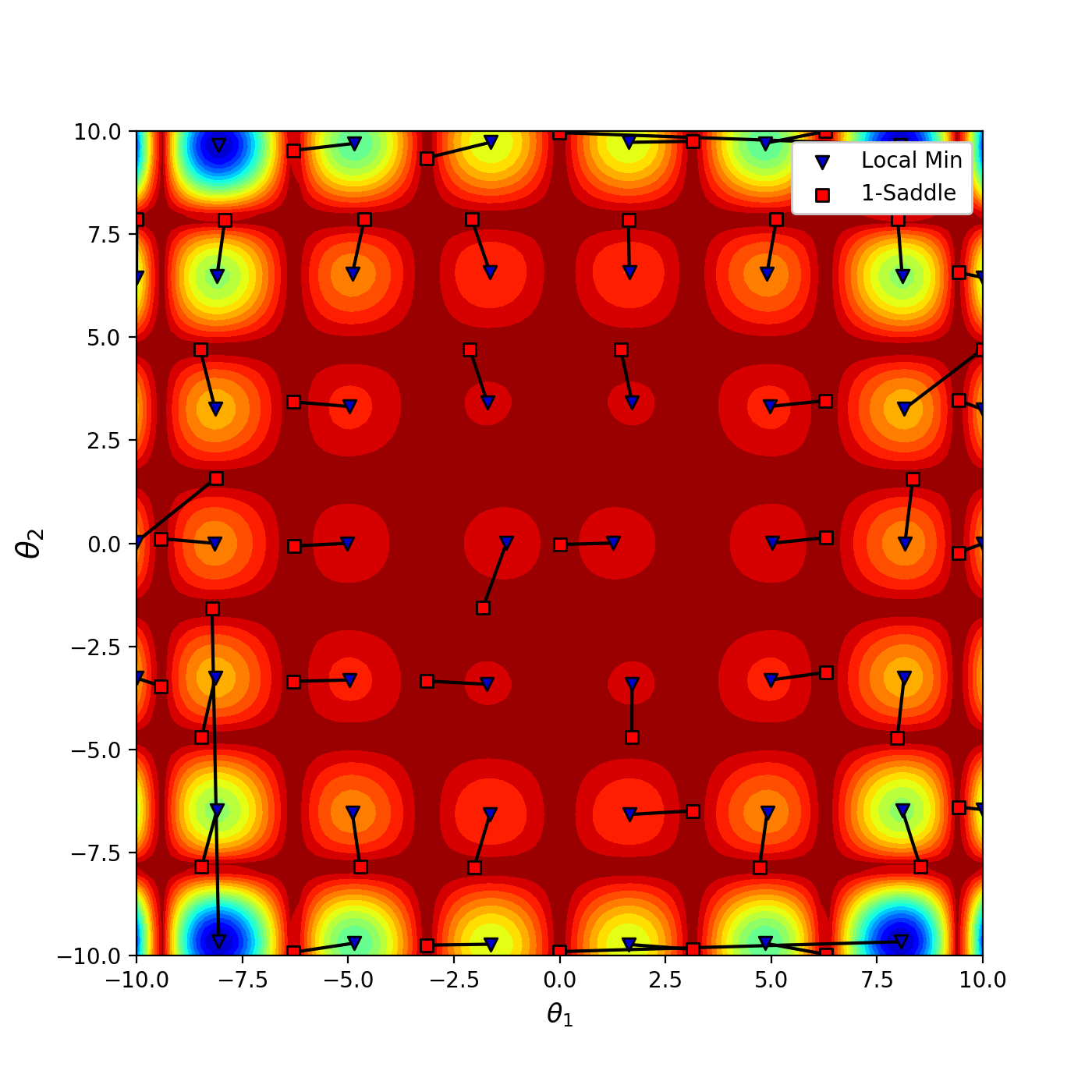}}\hfil 
\subfloat[Barcode of local minima.\label{fig:holderbarcodes}]{\includegraphics[width=0.33\linewidth]{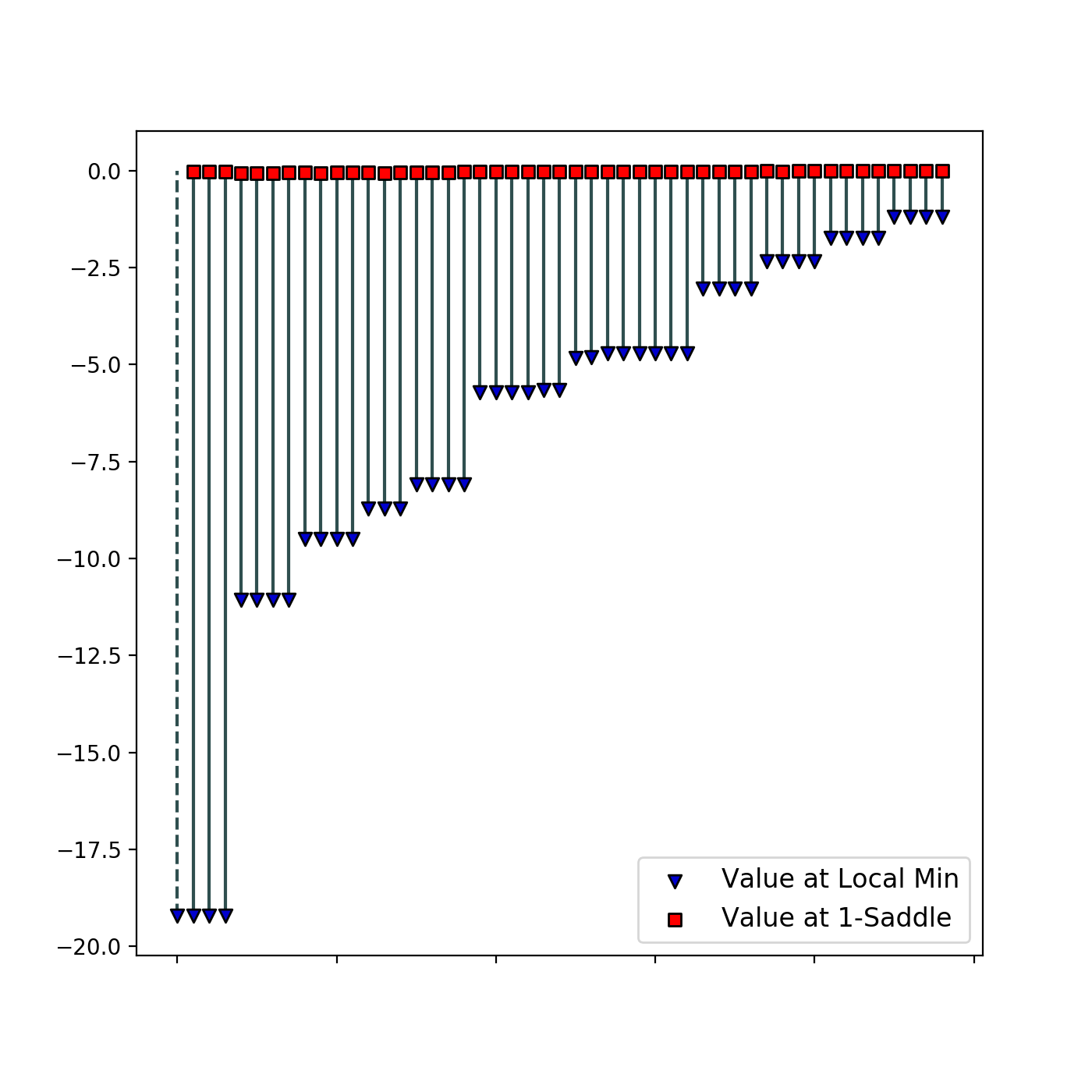}} 
\caption{HolderTable function, its minima-saddles correspondence and barcode computed by Algorithm \ref{algorithm-graph}}
\label{fig:holder-plot}
\end{figure}

\end{enumerate}

\textbf{Convergence of the Algorithm}\label{sec:converge}

In this subsection we prove and test empirically the convergence of our algorithm when the number of points $N$ used to construct barcodes tends to infinity. To compare two barcodes  we adopt  \textbf{Bottleneck distance} \cite{efrat2001geometry,viterbo:2018}, also known as Wasserstein-$\infty$ distance $\mathbb{W}_{\infty}$, on the corresponding persistence diagrams ($2$-dimensional point clouds of birth-death pairs):

\begin{equation}
  \mathbb{W}_{\infty}(\mathcal{D}, \mathcal{D}') := \inf_{\pi\in\Gamma(\mathcal{D},\mathcal{D}')}   \sup_{a\in \mathcal{D}\cup\Delta} 
  \lvert a - \pi(a) \rvert.
\end{equation}
Here $\Delta$ denotes the "diagonal" (pairs with birth equal to death)   and $\Gamma(\mathcal{D},\mathcal{D}')$ denotes the set of partial matchings between $\mathcal{D}$ and $\mathcal{D}'$ defined as bijections between $\mathcal{D}\cup\Delta$ and $\mathcal{D}'\cup\Delta$.

For a function $f:\Theta\rightarrow\mathbb{R}$ let $\mathcal{D}_{f}^{*}$ denote its barcode. 
Our algorithm uses finite number of $N$ randomly sampled points $\Theta_{N}=\{\theta_{1},\dots,\theta_{N}\}$ to compute approximation $\hat{\mathcal{D}}_{f}(\Theta_{N})$ of true barcode $\mathcal{D}_{f}^{*}$. Let $C$ denote the Lipschitz constant of $f$
\begin{proposition}\label{prop:convrg}
Let for any $\theta\in\Theta$ there exist $\theta_i\in\Theta_{N}$ such that $\lvert \theta-\theta_i\rvert<\varepsilon$. Then 
\begin{equation*}
     \mathbb{W}_{\infty}\big(\hat{\mathcal{D}}_{f}(\Theta_{N}), \mathcal{D}_{f}^{*})\big) <C\varepsilon 
\end{equation*}
\end{proposition}
\begin{proof}
This follows for example from (\cite{chazal2011scalar}, Lemma 1).
\end{proof}
Note that for typical sample $\Theta_{N}$  the maximal distance from  $\theta\in\Theta$ to $\Theta_{N}$  is $\sim N^{-\frac{1}{D}}$. 
It follows that $\hat{\mathcal{D}}_{f}(\Theta_{N})$ converges to $\mathcal{D}_{f}^{*}$ as $N\to \infty$ for any typical sequence of samples $\{\Theta_{N}\}$. And in particular \begin{equation}
     \mathbb{W}_{\infty}\big(\hat{\mathcal{D}}_{f}(\Theta_{N}), \hat{\mathcal{D}}_{f}(\Theta_{N}')\big)\to 0 \label{wpair}
\end{equation}
as $N\to \infty$, for any typical pair of sequences $\{\Theta_{N}\},\{\Theta_{N}'\}$.

\begin{figure}[h!]
    \centering
    \includegraphics[width=0.99\linewidth]{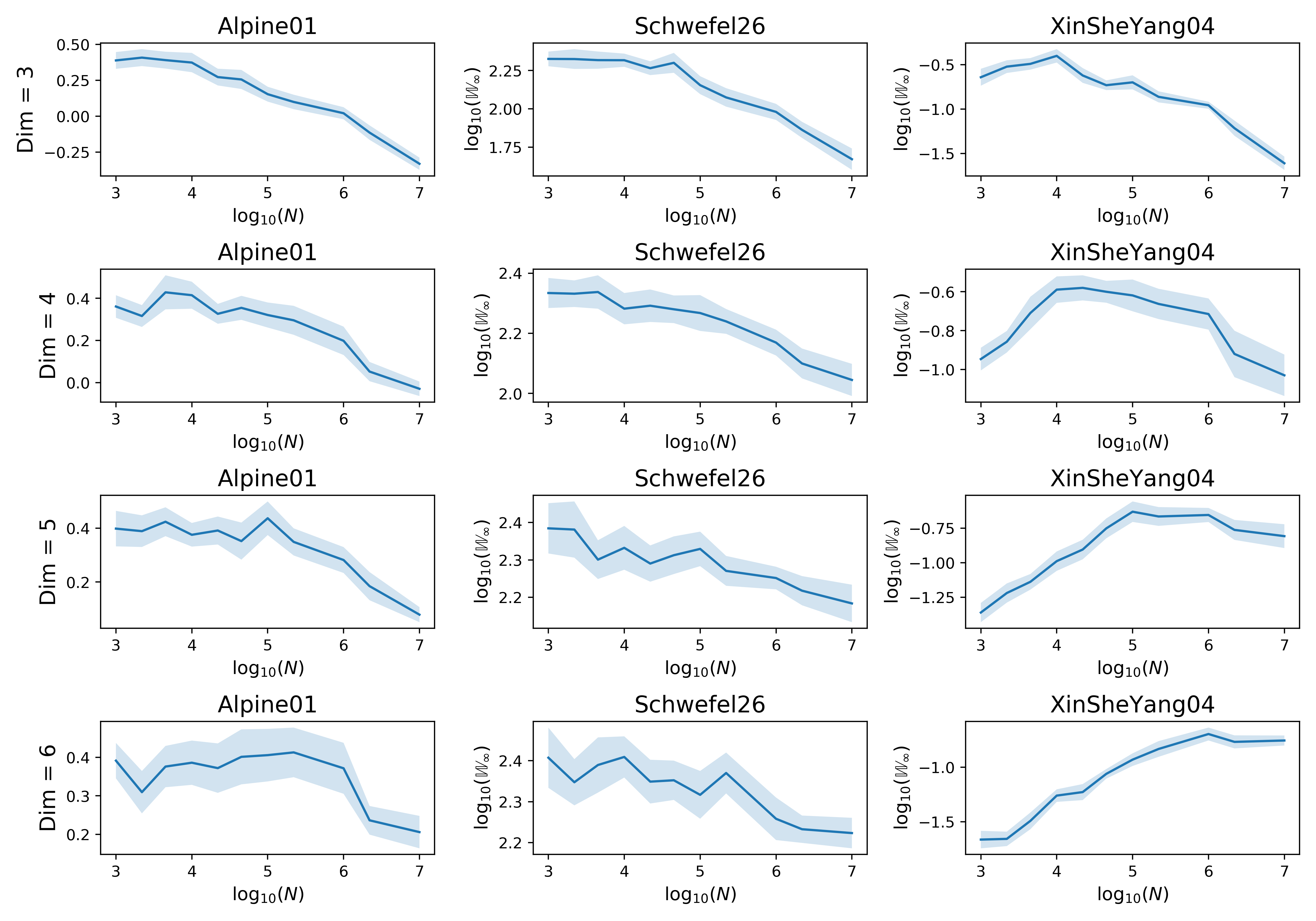}
    \caption{Decrease of logarithm of Bottleneck distance between pairs of persistent diagrams on samples of size $N$ for Alpine01, Schwefel26, XinSheYang04 benchmark functions in dimensions $D\in \{3,4,5,6\}$.}
    \label{fig:var-convergence}
\end{figure}

We have checked the condition (\ref{wpair}) on several test function from Global Optimization benchmark.
Each of the test function comes in a series of functions of arbitrary dimension. Even for small dimensions they are extremely complex (contain exponential number of local minima and saddle points). 
 For each function $f$ and dimension $D=3,4,5,6$, we consider various sample sizes $\log_{10} N\in [3, 3.5, 4, \dots, 7]$.
For every triplet $(f,D,N)$, we randomly sample $R=20$ pairs $\big((\Theta_{N})_{r},(\Theta_{N}')_{r}\big)$ of point clouds of size $N$. Next for each of $2R$ point clouds we compute the barcode by our algorithm and in each pair measure the bottleneck distance between the barcodes. Finally, we take the mean value
\begin{equation}
\mathbb{E}_{\Theta_{N},\Theta_{N}'}\mathbb{W}_{\infty}\big(\hat{\mathcal{D}}_{f}(\Theta_{N}), \hat{\mathcal{D}}_{f}(\Theta_{N}')\big)\approx \frac{1}{R}\sum_{r=1}^{R}\mathbb{W}_{\infty}\big(\hat{\mathcal{D}}_{f}((\Theta_{N})_{r}), \hat{\mathcal{D}}_{f}((\Theta_{N}')_{r})\big).\label{ETheta}
\end{equation}

 The considered functions are:

\begin{enumerate}
    \item \textbf{Alpine01} function $f:[-10,10]^{D}\rightarrow \mathbb{R}$ defined by
    \begin{equation}
    f(\theta_{1},\dots,\theta_{D})=\sum_{d=1}^{D}|\theta_{d}\sin \theta_{x_{d}}+0.1\theta_{d}|.
    \nonumber
    \label{alpine01}
    \end{equation}
    \item \textbf{Schwefel26} function $f:[-500,500]^{D}\rightarrow \mathbb{R}$ defined by:
    \begin{equation}
    f(\theta_{1},\dots,\theta_{D})=418.9829\cdot D-\sum_{d=1}^{D}\theta_{d}\sin(\sqrt{|\theta_{d}|}).
    \nonumber
    \label{schwefel26}
    \end{equation}
    \item \textbf{XinSheYang04} function $f:[-10,10]^{D}\rightarrow \mathbb{R}$ defined by:
    \begin{eqnarray}
     f(\theta_{1},\dots,\theta_{D})=\left[ \sum_{d=1}^{D} \sin^2(\theta_d) - e^{-\sum_{d=1}^{D} \theta_d^2} \right ] e^{-\sum_{d=1}^{D} \sin^2 \sqrt{ \lvert \theta_d \rvert }}
    \nonumber
    \label{XinSheYang04}
    \end{eqnarray}
\end{enumerate}

For every pair $(f,D)$ we sum up the dependence on cloud size $N$ in a form of a plot in the decimal logarithmic scale. 
We  observe  empirically that
for big enough $N$
\begin{equation*}
\log( \mathbb{W}_{\infty}\big(\hat{\mathcal{D}}_{f}(\Theta_{N}), \hat{\mathcal{D}}_{f}(\Theta_{N}')\big))\sim -\frac{1}{D}\log (N) \end{equation*}
as it is expected based on Proposition \ref{prop:convrg}. 

The results are summarized in Figure \ref{fig:var-convergence}. 

In Table 1 the running times for the Algorithm 1, including HNSW NN search,  on random samples of $10^6$ points, are compared with the running times for platform GUDHI on  $10^6$ points grid, for  Alpine01 and Schwefel26 benchmark functions in dimensions $D\in \{3,4,5,6,7\}$. 
\begin{table}[h!]
\resizebox{\textwidth}{!}{%
\begin{tabular}{|c|c|c|c|c|c|c|c|c|c|c|}
\hline
 & Alpine01 & Schwefel26 & Alpine01 & Schwefel26 & Alpine01 & Schwefel26 &  Alpine01 & Schwefel26 &  Alpine01 & Schwefel26 \\ \cline{2-11}
  & 3D & 3D  & 4D  & 4D  & 5D  & 5D & 6D  & 6D & 7D & 7D  \\ \hline
 Gudhi& 9 & 10 & 28 & 29 & 72 & 79 & 160 & 181 & 767 & 863 \\ \hline
 Alg 1& 12 & 12 & 14 & 14 & 16 & 16 & 18 & 18 & 22 & 21\\ \hline
\end{tabular}%
}
\caption{Running time in seconds on $10^6$ points, on Intel(R) Xeon(R) CPU E5-2698 v4 @2.20GHz}
\end{table}

\section{Topology of neural networks' loss functions}

\label{sec-nn}

\begin{figure}[h!]
    \includegraphics[width=0.94\linewidth]{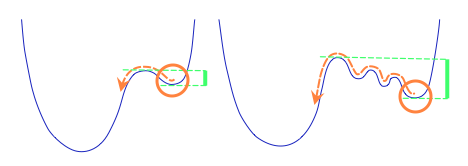}
    \caption{Barcodes quantify topological obstacles posed by local minima for gradient-based optimization. Here the two local minima locally look the same but clearly pose different obstacles for gradient-based learning. These obstacles are quantified by the lengths of green segments, which are associated with the local minima in the barcode.}
    \label{fig:obstacles}
\end{figure}

Clearly some minima can harm more than others gradient-based algorithms. For example two minima on  Figure \ref{fig:obstacles} locally look the same but pose different obstacles for gradient-based optimization. A gradient-based algorithm trajectory can have difficulty to escape from a vicinity of local minimum. To reach a point with lower loss, a path, coming from a vicinity of a given local minimum, has to climb up to points with higher loss. The minimal value of this higher loss penalty is precisely the loss  at 1-saddle associated with the minimum, from definition of barcode.    
The bigger the minimum's segment in the barcode  the more difficulty a gradient-based learning trajectory has escaping vicinity of the minimum in order to reach lower loss points. 




In this section we compute and analyse barcodes of small fully connected neural networks with up to three hidden layers.

For several architectures of the neural networks  many results on the loss surface are known (see e.g. \cite{kawaguchi2016deep,gori1992problem} and references therein). Different geometrical and topological properties of loss surfaces were studied in   \cite{cao2017flatness,yi2019positively,wideval,sharpM}.

We have analyzed neural networks that are small. However our method permits full exploration of the loss surface as opposed to stochastical exploration of higher-dimensional loss surfaces. Let us emphasize that even from practical point of view it is important to understand first the behavior of barcodes in simplest examples where all hyper-parameters optimization schemes can be easily turned off.

For every analysed neural network the loss function is the mean squared error for predicting (randomly selected) function $g:[-\pi, \pi]\rightarrow \mathbb{R}$ given by
$$g(x)=0.31 \cdot \sin(-x) - 0.72 \cdot \sin(-2x) - 0.21 \cdot \cos(x) + 0.89 \cdot \cos(2x)$$ plus $l_2-$regularization.
The error is computed for prediction on  uniformly distributed inputs $x\in \{-\pi + \frac{2\pi}{100}k \text{ | } k=0,1,\dots, 100 \}$.


\begin{figure}[h]
\centering
\subfloat[Barcode for ($2$) net]{\includegraphics[width=0.33\linewidth]{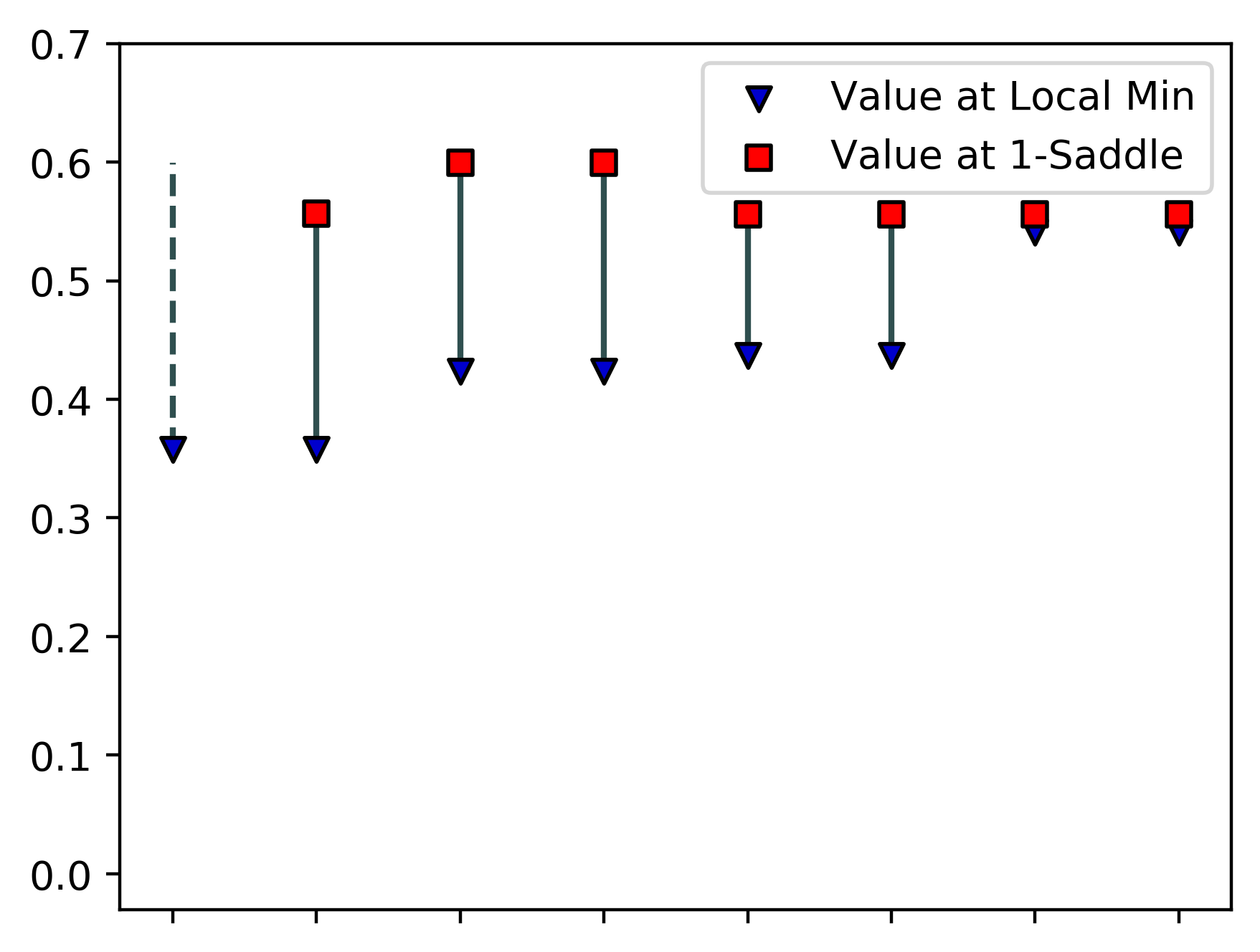}} \hfil
\subfloat[Barcode for  ($3$) net ]{\includegraphics[width=0.33\linewidth]{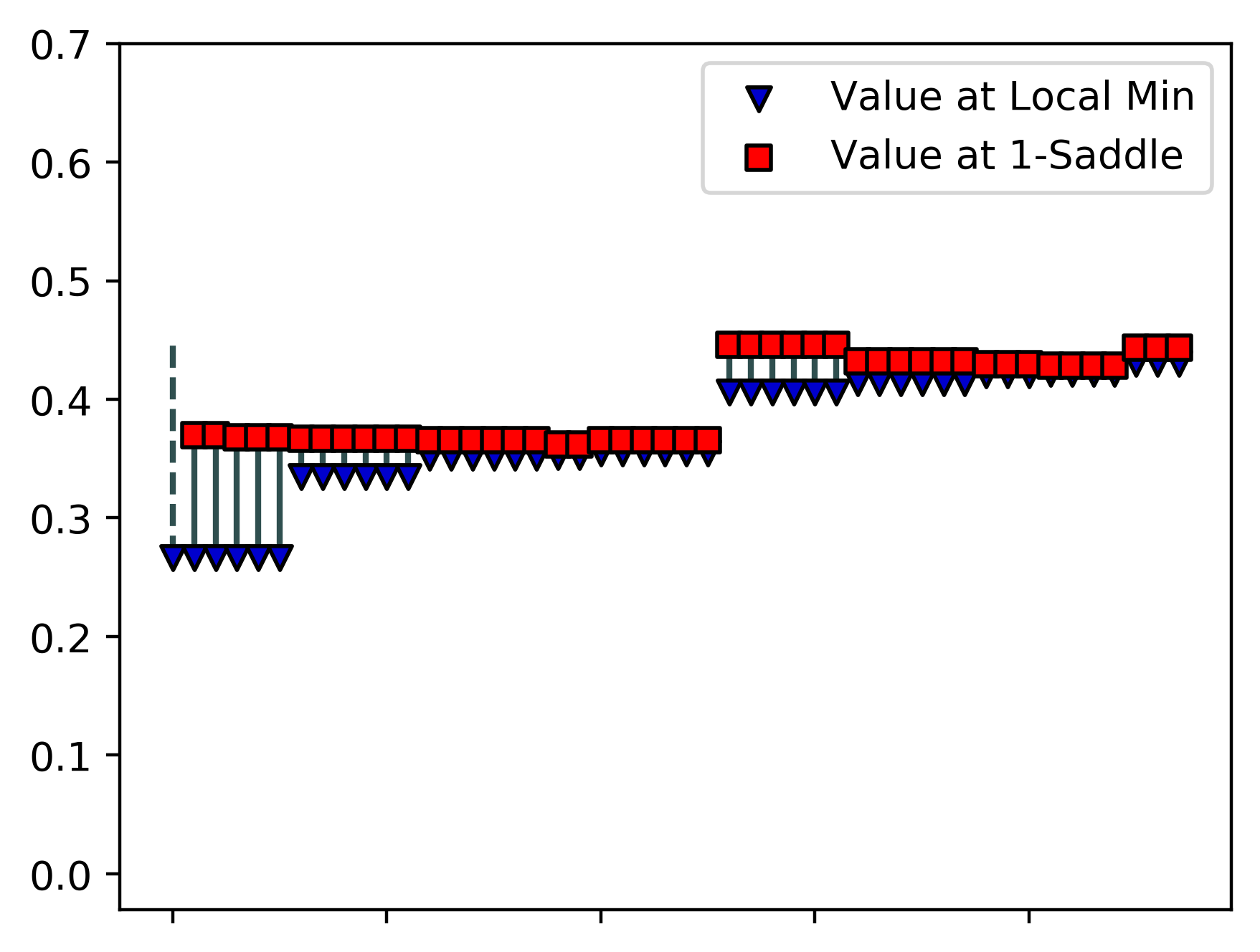}}\hfil

\subfloat[Barcode for ($2\times 2$) net]{\includegraphics[width=0.33\linewidth]{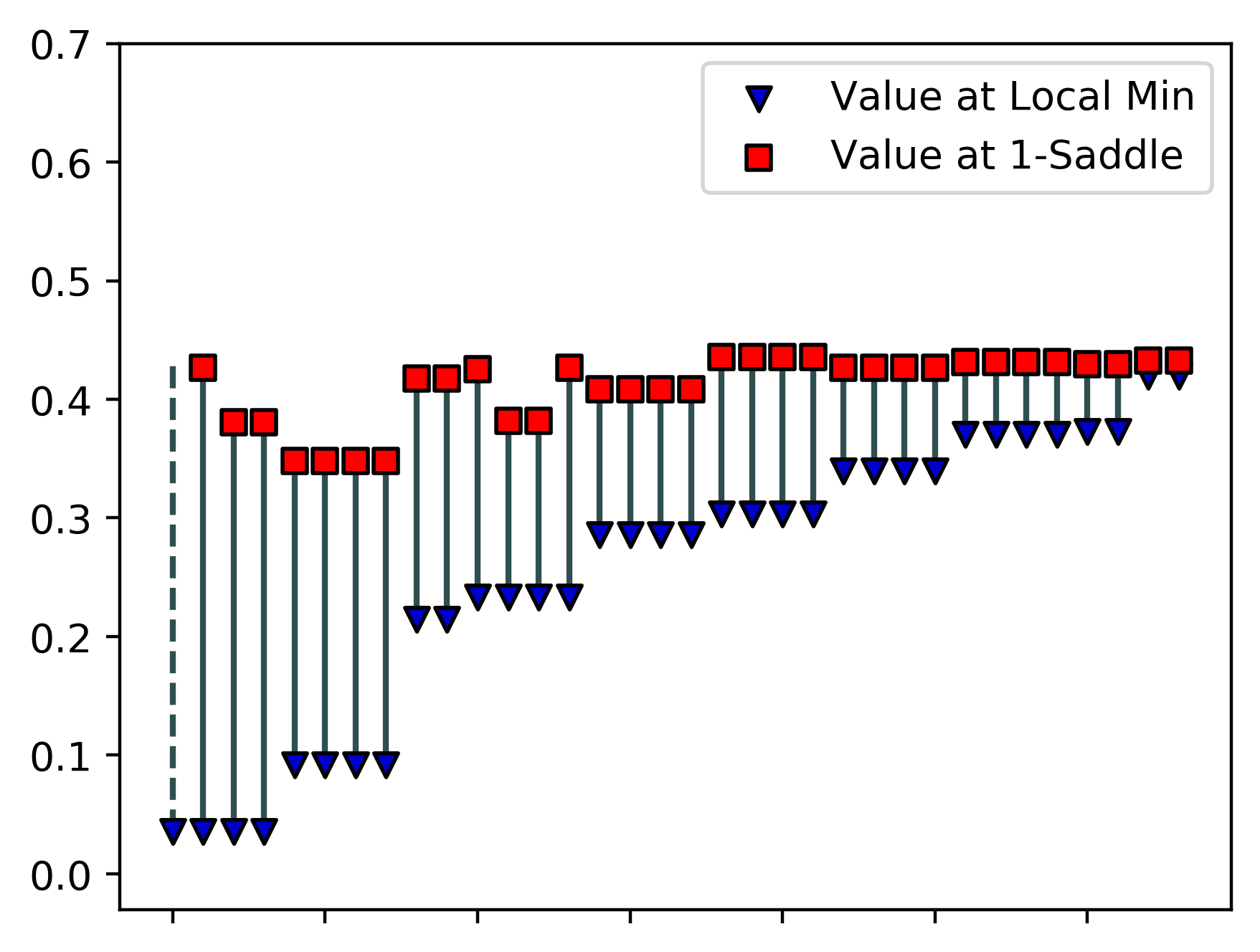}}\hfil
\subfloat[Barcode for ($3\times 2$) net]{\includegraphics[width=0.33\linewidth]{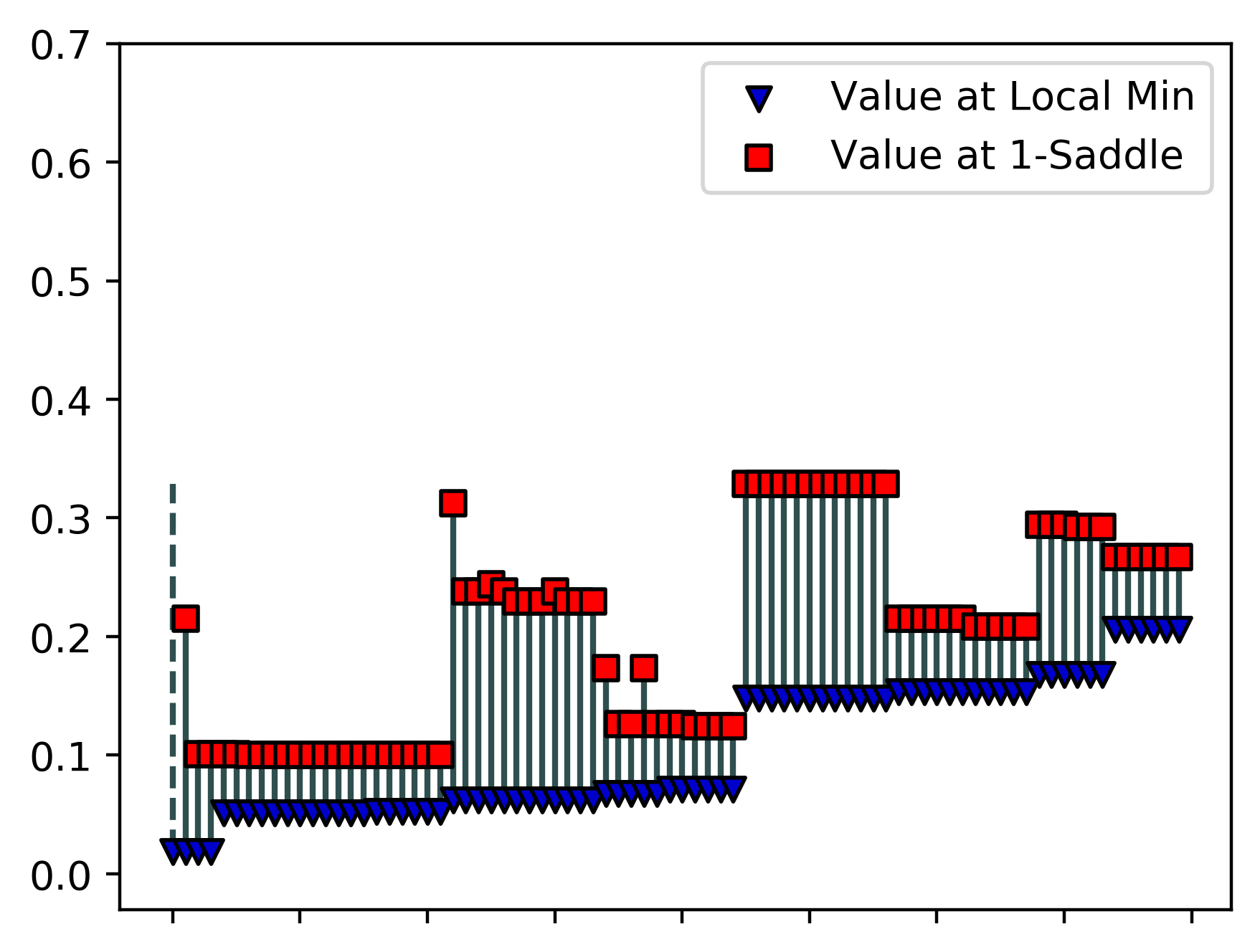}}\hfil
\subfloat[Barcode for ($3\times 3$) net ]{\includegraphics[width=0.33\linewidth]{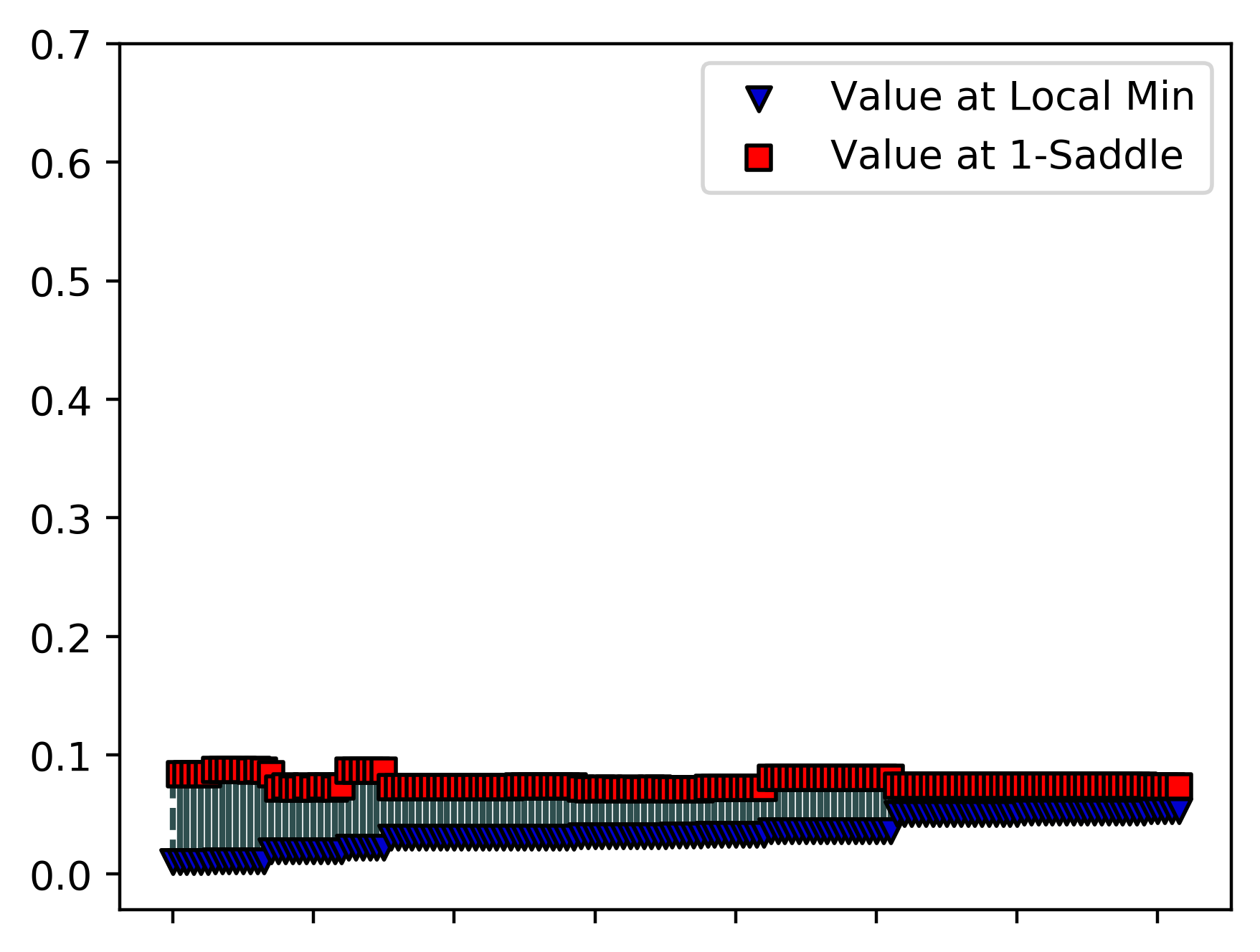}} \hfil

\subfloat[Barcode for ($2\times 2\times 2$) net]{\includegraphics[width=0.33\linewidth]{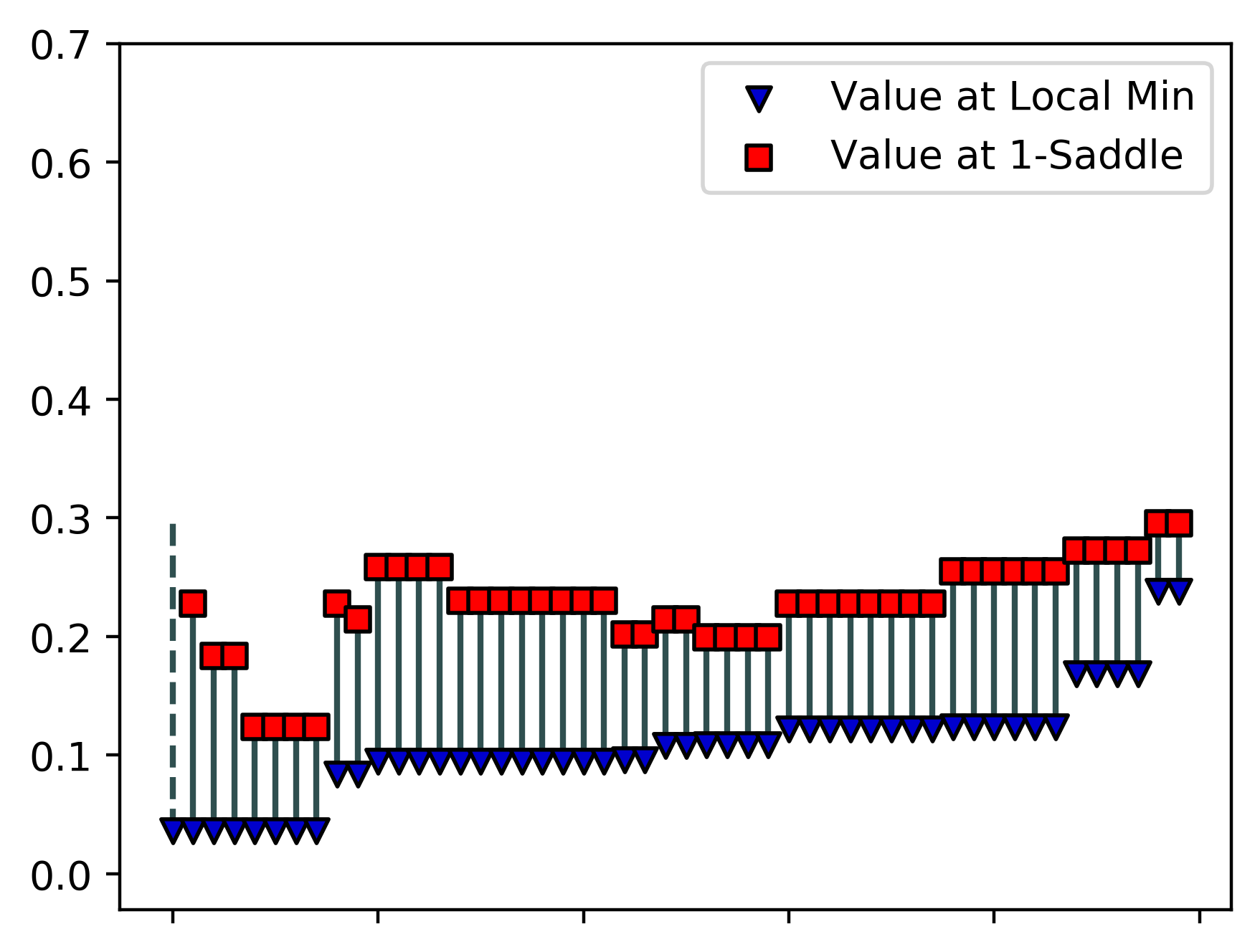}}\hfil   
\subfloat[Barcode for ($3\times 2\times 2$) net]{\includegraphics[width=0.33\linewidth]{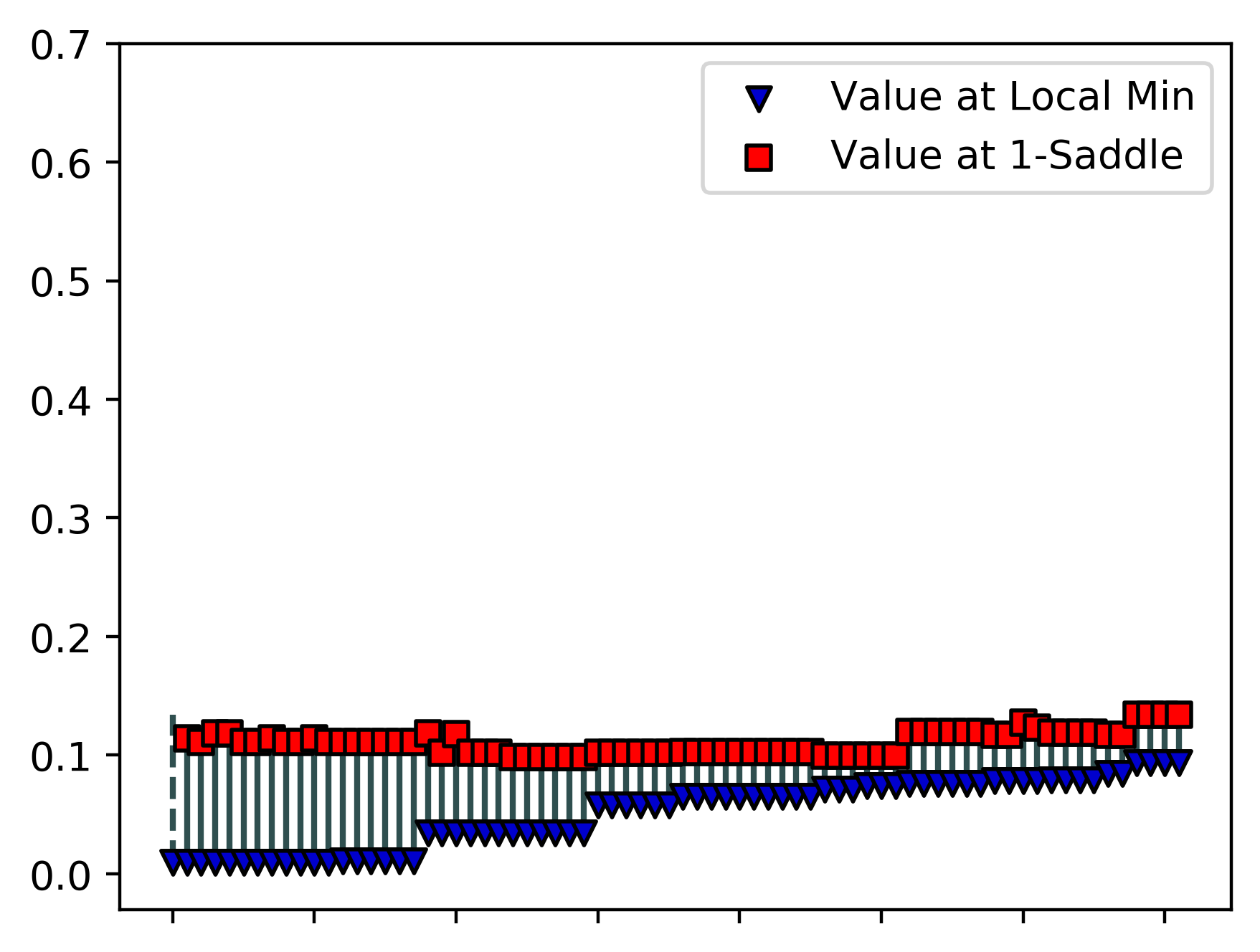}}
\caption{Barcodes of different neural network loss surfaces.}
\label{fig:network-barcodes}
\end{figure}

The neural networks considered were fully connected one-hidden layer with 2 and 3 neurons, two-hidden layers with 2x2, 3x2 and 3x3 neurons, and three hidden layers with 2x2x2 and 3x2x2 neurons with $ReLU$ activation. 
We have calculated the barcodes of the loss functions on the hyper-cubical sets $\Theta$ which were chosen based on the typical range of parameters of minima. The results are as shown in Figure \ref{fig:network-barcodes}.

We summarize our findings into two main observations:
\begin{enumerate}
    \item The barcodes are located in tiny lower part of the range of values; typically the maximum value of the function was around 200 and higher, and the saddles paired with minima lie well below $1$; 
    
    \item With the increase of the neural network depth and width the barcodes descend lower. 
\end{enumerate}

For example the upper bounds of barcodes of one-layer $(2)$ net are in range $[0.55,0.65]$, two-layer $(2\times 2)$ net in range $[0.35,0.45]$, and three-layer $(2\times 2\times2)$ net in range $[0.1,0.3]$.

\textbf{Implications for learning.}

All minima in the observed cases are located in low part of loss function's range and they descend lower as the depth and the width of neural network increases. 
The gradient flow trajectories always have minima as terminal points for all  but a subset of measure zero starting points.   
It follows that essentially any terminal point of gradient flow gives rather good solution to the regression problem. The precision of solution increases with increase of the neural network depth and width.

During learning, the gradient descent trajectory cannot get stuck at high local minima, since essentially all minima are located in a tiny low part of the function's range. There could perhaps exist some noise minima slightly higher but they  are easily escaped during learning since their barcode is low, which implies that there always exists an escape path with low penalty.

If the length of barcode's segment is small for any local minimum then  there always exists a small loss path to a lower minimum, this implies that gradient descent based optimization methods can in principle reach the lowest minimum  during learning. 

If the barcodes of all local minima are low this means also that any two such minima can be connected by a small loss path. This means that for any two local  minima there exists continuous low loss transformation between predictions of one minimum to predictions of another minimum, implying that their predictions are somehow  equivalent.

\textbf{Strategy for computing barcodes for deep neural networks.}
\label{subs:Dnn}

The exploration of loss surface using topological data analysis methods can be extended to deep neural networks. The strategy for deep neural networks is based on the proposition \ref{prop:Paths} above. The idea is to  act by  gradient descent on a path starting from the given minimum and going to a point with lower minimum. The deformation of the path under action of gradient flow is given by action of the component of the gradient which is orthogonal to the tangent direction of the path. The gradient component which is parallel to the path's tangent direction is absorbed into reparametrization of the path.  Then the formula (\ref{eq:Paths}) gives an estimate for the critical value of  "death" 1-saddle  corresponding to this minimum.

\textbf{Implications for generalization.} 

We have conducted some experiments that show that there exists a correlation between the height of barcodes of  groups of low  minima of same loss value and their generalization errors. Thus at least under certain mild conditions the lower the barcode for  minima with same loss value the better their generalization properties.

\section{Conclusion}
In this work we have introduced a methodology for analysing the graphs of functions, in particular, loss surfaces of neural networks. The methodology is based on computing topological invariants called barcodes.

To compute barcodes we used a graph-based construction which approximates the function. Then we apply the algorithm we developed to compute the function's barcodes of local minima. Our experimental results of computing barcodes for small neural networks lead to two principal observations.

First all barcodes sit in a tiny lower part of the total function's range. 
 Secondly the barcodes descend lower as the depth and width of neural network increases. From the practical point of view, this means that gradient descent optimization cannot get stuck in high local minima, and it is also not difficult to get from one local minimum to another (with smaller value) during learning.

The method that we developed has several further research directions. Although we tested the method on small neural networks, it is possible to apply it to large-scale modern neural networks such as convolutional networks (i.e. ResNet, VGG, AlexNet, U-Net, see \cite{alom2018history}) for image-processing based tasks. However, in this case the graph-based approximation that we use requires wise choice of representative graph vertices as, dense filling of area by points is computationally intractable.  There are clearly also connections, deserving further investigation, between the  barcodes of local minima and optimal learning rates or  the rates of convergence during learning.

This work was partially supported by RFBR grant 21-51-12005 NNIO\_a




\appendix
\section*{Appendix}

\section{Gradient Morse complex}\label{subsMorse}
The gradient Morse complex $(C_*,\partial_{*})$, is defined as follows.  
For generic $f$ the   critical points $p_{\alpha}$,  $df\mid_{T_{p_{\alpha}}}=0$, are isolated. Near each critical point $p_{\alpha}$ $f$ can be written as $f=\sum_{l=1}^{j}-(x^{l})^{2}+\sum_{l=j}^{n}(x^{l})^{2}$ in some local coordinates.
The index of the critical point is defined as the dimension of the set of downward pointing directions at that point, or of the negative subspace  of the Hessian: $$ \textrm{index}(p_{\alpha})=j$$

 Then define $$C_{j}=\oplus_{\textrm{index}(p_{\alpha})=j}\left[p_{\alpha},\textrm{or}(T_{p_{\alpha}}^{-})\right]$$ where $\textrm{or}$ is an orientation on a negative subspace  $T_{p_{\alpha}}=T_{p_{\alpha}}^{-}\oplus T_{p_{\alpha}}^{+}$ of the Hessian $\partial^{2}f$. 

Let 
$$ \mathcal{M}(p_{\alpha},p_{\beta})=\left\{ \gamma:\mathbb{R}\rightarrow M^{n}\mid\right.
\left.\dot{\gamma}=-(\textrm{grad}_{g}f)(\gamma(t)),\lim_{t\rightarrow-\infty}=p_{\alpha},\lim_{t\rightarrow+\infty}=p_{\beta}\right\} /\mathbb{R}
$$
be the set of gradient trajectories connecting  critical points $p_\alpha$
and $p_\beta$, where the natural action of $\mathbb{R}$ is by the shift $\gamma(t)\mapsto\gamma(t+\tau)$.

If  $\textrm{index}(p_{\beta})= \textrm{index}(p_{\alpha})-1$ then generically the set  $\mathcal{M}(p_{\alpha},p_{\beta})$ is finite. 
Let $$\#\mathcal{M}(\left[p_{\alpha},\textrm{or}\right],\left[p_{\beta},\textrm{or}\right])$$ denote in this case the number of the trajectories, counted with signs taking into account a choice of orientation, between critical points $p_\alpha$ and $p_\beta$. 

The linear operator $\partial_{j}$ is defined by  \begin{equation*}
\partial_{j}\left[p_{\alpha},\textrm{or}\right]=\sum_{\textrm{index}(p_{\beta})=j-1}\left[p_{\beta},\textrm{or}\right]\#\mathcal{M}(p_{\alpha},p_{\beta})
\end{equation*}


The description of the critical points on  manifold $\Theta$ with nonempty boundary ${\partial \Theta}$ is modified slightly in the following way. A connected component of sublevel set is  born also at a local minimum of restriction of $f$ to the boundary $f\left. \right|_{\partial \Theta}$, if $\text{grad}f$ is pointed inside manifold $\Theta$. The merging of two connected components can also happen at  1-saddle of $f\left. \right|_{\partial \Theta}$, if $\text{grad}f$ is pointed inside $\Theta$. When we speak about minima and 1-saddles, this  also means such critical points of $f\left. \right|_{\partial \Theta}$. Similarly the set of generators of index $j$ chains in Morse complex includes index $j$ critical points of $f\left. \right|_{\partial \Theta}$ with $\text{grad}f$  pointed inside $\Theta$. The differential is also modified similarly to take into account  trajectories involving such critical points.

\section{Proof of the theorem \ref{theorem1}}\label{proofTh1}

\noindent
\textbf{Theorem.} (\cite{B94}, Section 2)
\emph{Any $\mathbb{R}-$filtered chain complex  $C_{*}$  over field $\mathtt{k}$
 can be brought by a linear transformation preserving
the $\mathbb{R}-$filtration to ``canonical form'', a canonically defined direct
sum of indecomposable $\mathbb{R}-$filtered complexes of two types: \begin{itemize}
    \item 
1-dimensional $\mathbb{R}-$filtered
complex with trivial differential, $\partial\tilde{e}_{i}^{(j)}=0$
, $\left\langle \tilde{e}_{i}^{(j)}\right\rangle =F_{\leq r}$, $r\in\mathbb{R}$,
\item 2-dimensional $\mathbb{R}-$filtered complex with trivial homology
$\partial\tilde{e}_{i_{2}}^{(j+1)}=\tilde{e}_{i_{1}}^{(j)}$, $\left\langle \tilde{e}_{i_{1}}^{(j)}\right\rangle =F_{\leq s_{1}}$, $\left\langle \tilde{e}_{i_{1}}^{(j)},\tilde{e}_{i_{2}}^{(j+1)}\right\rangle =F_{\leq s_{2}}$,
$s_{1},\,s_{2}\in\mathbb{R}$.\end{itemize}  The resulting canonical form is unique.}

\begin{proof} (\cite{B94}, Section 2)

Let $\{e_{i}^{(n)}\}$ be a basis in the vector spaces $C_{n}$ compatible with the filtration, so that each subspace $F_{r}C_{n}$ is the span $\left\langle e_{1}^{(n)},\ldots,e_{i_{r}}^{(n)}\right\rangle $. Notice that the filtration defines the natural order on the set of basis elements.

Let   $\partial e_{l}^{(n)}$ have the required form for $n=j$ and $l\leq i$, or $n<j$ and all $l$. I.e. either $\partial e_{l}^{(n)}=0$ or $\partial e_{l}^{(n)}=e_{m(l)}^{(n-1)}$, where $m(l)\neq m(l')$ for $l\neq l'$. 

Let $$\partial e_{i+1}^{(j)}=\sum_{k}e_{k}^{(j-1)}\alpha_{k}.$$ Let's move all the terms with $e_{k}^{(j-1)}=\partial e_{q}^{j}$, $q\leq i$, from the right to the left side. We get $$\partial(e_{i+1}^{(j)}-\sum_{q\leq i}e_{q}^{(j)}\alpha_{k(q)})=\sum_{k}e_{k}^{(j-1)}\beta_{k}$$ If $\beta_{k}=0$ for all $k$, then define $$\tilde{e}_{i+1}^{(j)}=e_{i+1}^{(j)}-\sum_{q\leq i}e_{q}^{(j)}\alpha_{k(q)},$$ so that $$\partial\tilde{e}_{i+1}^{(j)}=0,$$ and $\partial e_{l}^{(n)}$ has the required form for $l\leq i+1$ and $n=j$, and for $n<j$ and all $l$.

Otherwise let $k_{0}$ be the maximal $k$ with $\beta_{k}\neq 0$. Then $$\partial(e_{i+1}^{(j)}-\sum_{q\leq i}e_{q}^{(j)}\alpha_{k(q)})=e_{k_{0}}^{(j-1)}\beta_{k_{0}}+\sum_{k<k_{0}}e_{k}^{(j-1)}\beta_{k},$$
$\beta_{k_{0}}\neq 0.$ 
Define $$\tilde{e}_{i+1}^{(j)}=\left(e_{i+1}^{(j)}-\sum_{q\leq i}e_{q}^{(j)}\alpha_{k(q)}\right)/\beta_{k_{0}},$$
$$\tilde{e}_{k_{0}}^{(j-1)}=e_{k_{0}}^{(j-1)}+\sum_{k<k_{0}}e_{k}^{(j-1)}\beta_{k}/\beta_{k_{0}}.$$ 
Then $$\partial\tilde{e}_{i+1}^{(j)}=\tilde{e}_{k_{0}}^{(j-1)}$$ and for $n=j$ and $l\leq i+1$, or $n<j$ and all $l$, $\partial e_{l}^{(n)}$ has the required form. If the complex has been reduced to "canonical form" on subcomplex  $\oplus_{n \leq j} C_n$, then reduce similarly $\partial e_{1}^{(j+1)}$ and so on.

Uniqueness of the canonical form follows essentially from the uniqueness at each previous step.
Let $\left\{ a_{i}^{(j)}\right\} $, $\left\{ b_{i}^{(j)}=\sum_{k\leq i}a_{k}^{(j)}\alpha_{k}\right\}$
be two bases of $C_{*}$ for two different canonical forms. Assume that for
all indexes $p<j$ and all $n$, and $p=j$ and $n\leq i$ the canonical
forms agree. Let $\partial a_{i+1}^{(j)}=a_{m}^{(j-1)}$ and $\partial b_{i+1}^{(j)}=b_{l}^{(j-1)}$
with $m>l$, $a_{m}^{(j-1)}$ is
not in the filtration subspace corresponding to $b_{l}^{(j-1)}$.

It follows that 
$$
\partial\left(\sum_{k\leq i+1}a_{k}^{(j)}\alpha_{k}\right)=\sum_{n\leq l}a_{n}^{(j-1)}\beta_{n},
$$
where $\alpha_{i+1}\neq0$, $\beta_{l}\neq0$. Therefore 
$$
\partial a_{i+1}^{(j)}=\sum_{n\leq l}a_{n}^{(j-1)}\beta_{n}/\alpha_{i+1}-\sum_{k\leq i}\partial a_{k}^{(j)}\alpha_{k}/\alpha_{i+1}.
$$

On the other hand $\partial a_{i+1}^{(j)}=a_{m}^{(j-1)}$, with $m>l$,
and $\partial a_{k}^{(j)}$ for $k\leq i$ are either zero or some
basis elements $a_{n}^{(j-1)}$ different from $a_{m}^{(j-1)}$. This gives a contradiction.

Similarly if $\partial b_{i+1}^{(j)}=0$, then $$\partial a_{i+1}^{(j)}=-\sum_{k\leq i}\partial a_{k}^{(j)}\alpha_{k}/\alpha_{i+1}$$ which again gives a contradiction by the same arguments. This proves the uniqueness of the canonical form.
\end{proof}

\begin{remark} The barcode of $\mathbb{R}-$filtered chain complex consists of segments representing the indecomposable
$\mathbb{R}-$filtered chain complexes, see Definition \nolinebreak\ref{defin2} in Section \ref{sec:2ndDef}.
There is the standard lexicographic order on a set of such segments.
The direct sum from the theorem statement  is the standard
 ``direct sum over a set'' vector space.  \end{remark}



\bibliography{references.bib}

\bibliographystyle{elsarticle-num}



\end{document}